\documentclass[conference]{IEEEtran}
\IEEEoverridecommandlockouts
\usepackage{subfigure}
\usepackage{multirow}
\usepackage{threeparttable}
\usepackage{booktabs}
\usepackage{cite}
\usepackage{amsmath,amssymb,amsfonts}
\usepackage{algorithmic}
\usepackage{textcomp}
\usepackage{xcolor}
\usepackage{color, soul, framed}
\usepackage{verbatimbox}
\usepackage{enumitem}%
\usepackage{amsmath}
\usepackage{amssymb}
\usepackage[ruled]{algorithm2e}
\usepackage{etoolbox}
\makeatletter
\patchcmd{\@makecaption}
  {\scshape}
  {}
  {}
  {}
\makeatletter
\patchcmd{\@makecaption}
  {\\}
  {.\ }
  {}
  {}
\makeatother

\newcommand{\todo}[1]{#1}

\newcommand{\ie}[0]{\textit{i.e.}}
\newcommand{\eg}[0]{\textit{e.g.}}

\newtheorem{MyDef}{Definition}
\newcommand{\tabincell}[2]{
\begin{tabular}{@{}#1@{}}#2\end{tabular}
}

\makeatletter
\IEEEtriggercmd{\reset@font\normalfont\fontsize{7.9pt}{8.40pt}\selectfont}
\makeatother
\IEEEtriggeratref{1}

\ifCLASSINFOpdf
   \usepackage[pdftex]{graphicx}
   \graphicspath{{../pdf/}{../jpeg/}}
\else
\fi
\usepackage{amsmath}
\usepackage{algorithmic}
\begin{document}
%
\title{HybridGNN: Learning Hybrid Representation for Recommendation in Multiplex Heterogeneous Networks
}


\author{Tiankai Gu\textsuperscript{\rm $\dagger$*}\thanks{* Equal contribution},
Chaokun Wang\textsuperscript{\rm $\dagger$},
Cheng Wu\textsuperscript{\rm $\dagger$*}, 
Jingcao Xu\textsuperscript{\rm $\dagger$},
Yunkai Lou\textsuperscript{\rm $\dagger$},\\
Changping Wang\textsuperscript{\rm $\ddagger$},
Kai Xu\textsuperscript{\rm $\ddagger$},
Can Ye\textsuperscript{\rm $\ddagger$},
Yang Song\textsuperscript{\rm $\ddagger$}\\
\textsuperscript{\rm $\dagger$}School of Software, Tsinghua University, Beijing 100084, China\\
\textsuperscript{\rm $\ddagger$}Kuaishou Inc., Beijing 100000, China\\
\textsuperscript{\rm $\dagger$}\{gtk18, c-wu19, xjc20, louyk18\}@mails.tsinghua.edu.cn, chaokun@tsinghua.edu.cn, \\ \textsuperscript{\rm $\ddagger$}\{wangchangping, xukai, yecan, yangsong\}@kuaishou.com
}

\maketitle

\begin{abstract}
Recently, graph neural networks have shown the superiority of modeling the complex topological structures in heterogeneous network-based recommender systems. 
Due to the diverse interactions among nodes and abundant semantics emerging from diverse types of nodes and edges, there is a bursting research interest in learning expressive node representations in multiplex heterogeneous networks.
One of the most important tasks in recommender systems is to predict the potential connection between two nodes under a specific edge type (\ie, relationship). 
Although existing studies utilize explicit metapaths to aggregate neighbors, practically they only consider intra-relationship metapaths and thus fail to leverage the potential uplift by \textit{inter-relationship} information.
Moreover, it is not always straightforward to exploit inter-relationship metapaths comprehensively under diverse relationships, especially with the increasing number of node and edge types. 
In addition, contributions of different relationships between two nodes are difficult to measure.
To address the challenges, we propose \textit{HybridGNN}, an end-to-end GNN model with hybrid aggregation flows and hierarchical attentions to fully utilize the heterogeneity in the multiplex scenarios.
Specifically, \textit{HybridGNN} applies a randomized inter-relationship exploration module to exploit the multiplexity property among different relationships. 
Then, our model leverages hybrid aggregation flows under intra-relationship metapaths and randomized exploration to learn the rich semantics.
To explore the importance of different aggregation flow and take advantage of the multiplexity property, we bring forward a novel hierarchical attention module which leverages both metapath-level attention and relationship-level attention.
Extensive experimental results on five real-world datasets suggest that \textit{HybridGNN} achieves the best performance compared to several state-of-the-art baselines ($p<0.01, t$-test) with statistical significance.
\end{abstract}



%
\IEEEpeerreviewmaketitle

\section{Introduction}
\begin{figure}[ht]
\centering
\includegraphics[width=1.0\linewidth]{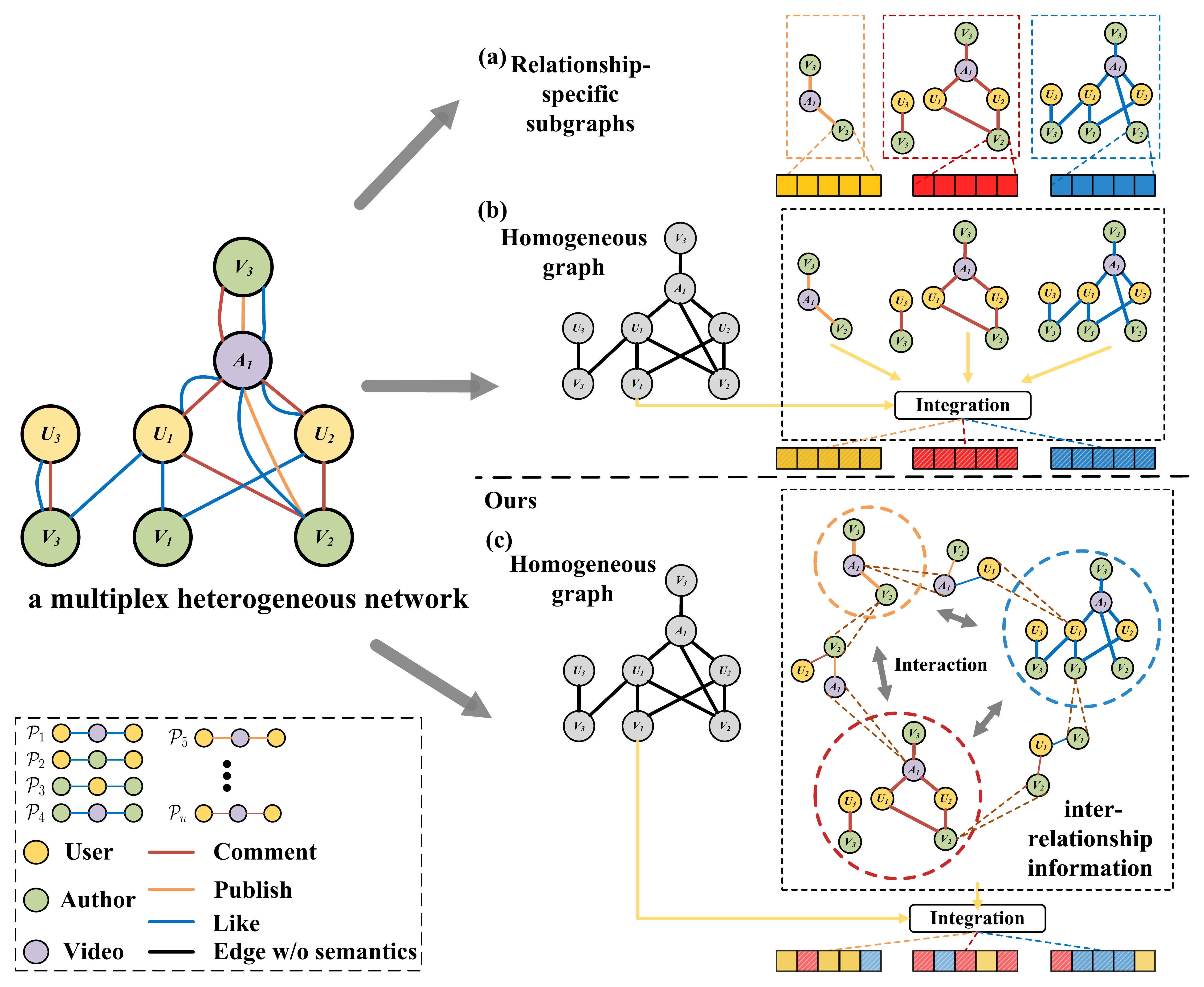} 
\caption{An illustration of different relationship-specific representation learning methods. (a) Learning node embedding through relationship-specific subgraphs by treating each subgraph independently. (b) Applying an interaction module with a base embedding across all relationships to learn node embeddings. It ignores inter-relationships and hence loses significant semantics and fails to fully utilize the multiplexity property. (c) Our method learns relationship-specific embeddings by considering both inter-relationships and topological structures.}
\label{fig:intro_example}
\end{figure}
In recent years, exploiting user interests through network embedding has emerged as a trending research topic in the field of recommender systems\todo{\cite{tan2021multi,zhao2021variational,yin2019social}}.
Traditional graph embedding methods such as DeepWalk \cite {perozzi2014deepwalk}, node2vec \cite{grover2016node2vec} and graph neural networks like GCN \cite{kipf2017semi}, GraphSage \cite{hamilton2017inductive} have achieved good performances in many tasks such as node classification, node clustering and link prediction.
Recently, heterogeneous graphs \cite{chang2015heterogeneous}, which is also known as heterogeneous information networks (HINs) \cite{su2009survey, shi2018heterogeneous} or heterogeneous networks, have been extensively studied to exploit rich semantics from multiple node and edge types. 
As a metapath in HINs represents a composite relationship implying specific semantic information, a variety of metapath-based models have been proposed to learn node representations in this paradigm \cite{chang2015heterogeneous, jin2020efficient, fu2020magnn, fan2019metapath, cen2019representation, zhang2018scalable}.  

In many real-world applications, multiplexity \cite{zhang2018scalable, wang2021fastsgg} is a common property in HINs, where there are usually multiple types of edges between two nodes. 
For example, in Kuaishou,  one of the largest online short-video platforms, there are three kinds of node types (\ie, author, user and video).
Each pair of nodes can be connected under multiple kinds of relationships (\eg, comment, publish and like).
A user may elect to comment and like a video simultaneously.
Moreover, connections between users and items in e-commerce systems have multiple types like page view, purchasing and adding to favorites~\todo{\cite{zhu2019joint, song2020poisonrec}}. 
In social networks, interactions between two persons are also multifaceted \cite{zhang2018scalable}, such as friendship relationship and money transferring relationship.

Making recommendations in multiplex scenarios is not an easy task. One of the key challenges of fully capturing the multiplexity property is to characterize  abundant  interactions  among  multiple types.
As shown in Fig.~\ref{fig:intro_example} (a), an intuitive idea for solving relationship-specific node representations is to apply methods on homogeneous graphs (\eg, GCN \cite{kipf2017semi}) to node embedding learning on relationship-specific subgraphs. 
However, those approaches fail to integrate local views from different types of subgraphs together.
Each subgraph is independently treated, which may cause information loss problems. 
Although existing works \cite{zhang2018scalable, cen2019representation} study relationship-specific representation of multiplex graphs, they still face the so-called \todo{``}heterogeneity deficiency" problem, which means lacking the ability to fully exploit and integrate both explicit and implicit semantics of the heterogeneous networks. 
For example, methods such as MNE \cite{zhang2018scalable}, as shown in Fig.~\ref{fig:intro_example} (b), only use a common embedding for two connected nodes across all edge types to leverage semantic correlation over different relationships. These methods are hard to fully exploit heterogeneity since cross-subgraph (\ie, inter-relationship) information and diversity of node types are ignored. 
\todo{Other methods like GATNE\cite{cen2019representation} and MAGNN\cite{fu2020magnn} explore more on correlations among different types of nodes or relationships. The former method adopts self-attention mechanisms to capture the interactions of relationships but it merely aggregates randomly sampled neighbors, which loses complex interactive information between nodes. The latter utilizes metapaths to model the heterogeneous node interactions and regards the aggregation of metapath instances as the embedding of the target node, but it is only designed for heterogeneous networks and neglects the relationship multiplexity. Besides, all of these methods intrinsically treat the whole graph as several relationship-specific subgraphs,
losing the graph's global properties to some extent.}

In summary, there are two major challenges in multiplex heterogeneous networks as following:

\begin{itemize}[noitemsep,topsep=0pt]
    \item \textit{Inter-relationship Exploration}. Although intra-relation-ship metapaths help understand user preferences under a relationship-specific subgraph, dependencies among different relationships remain unexplored.
    Moreover, only aggregating information from predefined metapaths may cause information loss, while enumerating all meaningful intra-relationship metapaths and inter-relationship metapaths is costly.
    To the best of our knowledge, most existing studies only specify several predefined metapaths and neglect inter-relationships to learn relationship-specific node representations.
    \item \textit{Hybrid Representation}. In multiplex heterogeneous networks, recommendations are not to simply predict potential connections, but to further predict connections under one specific relationship. Hence, it is necessary to learn different representations of a node under different relationships. However, although many works adopt neighbors aggregation to gather comparatively integrated information, they fail to notice the diversity of aggregation introduced by metapath schemes. Moreover, interactions of these latent metapath schemes are neglected. We argue that aggregation is of significance for relationship-specific representations in multiplex heterogeneous networks, and it is not comprehensively utilized in previous studies.
\end{itemize}

In order to tackle the above challenges, we propose \textit{HybridGNN}, a novel end-to-end GNN model to capture and learn different behaviors in multiplex heterogeneous networks for recommendation, as shown in Fig.~\ref{fig:intro_example} (c).
Firstly, \textit{HybridGNN} has a randomized inter-relationship exploration module to efficiently sample neighbors across relationship-specific subgraphs with a certain probability, \todo{which compensates the loss of global information by previous works.}
Secondly, \todo{compared with GATNE and MAGNN, \textit{HybridGNN} adopts hybrid aggregation flows to learn node embeddings via both intra-relationship and inter-relationship neighbors, with an aim to fully incorporate informative messages from heterogeneous networks.}
Thirdly, \todo{motivated by attention mechanisms \cite{wang2019hierarchical}}, \textit{HybridGNN} employs a hierarchical attention module, which mainly consists of two attention components: (a) the metapath-level attention to distinguish the importance of heterogeneous message passing from metapath-guided and randomized exploration guided neighbors, and (b) the relationship-level attention to comprehensively learn the latent connections among diverse relationships. 
Finally, \textit{HybridGNN} learns relationship-specific node embeddings to make recommendations.


The main contributions of this paper are as follows.
\begin{enumerate}[noitemsep,topsep=0pt]
\item We propose several novel concepts (Section \ref{sect:pre}) under the scenario of multiplex heterogeneous networks, based on which we derive a novel GNN model named \textit{HybridGNN}  (Section \ref{sect:overview}) to solve the recommendation issue. Our model uses hybrid aggregation flows to learn rich semantics of nodes (Section \ref{sect:intra}) and further uses hierarchical attention to fully utilize the heterogeneity in the multiplex heterogeneous network (Section \ref{sect:hie}). 
\item We design a randomized inter-relationship exploration approach for multiplex heterogeneous networks (Section \ref{sect:random}).
To the best of our knowledge, it is the first step toward incorporating inter-relationship information for learning relationship-specific node representations.
\item We conduct extensive experiments on five real-world datasets (Section \ref{sect:exp}). The experimental results demonstrate the superiority of our \textit{HybridGNN} model and illustrate a significantly statistical improvement compared to the state-of-the-art models ($p<0.01$, $t$-test). Moreover, the uplift of utilizing inter-relationships is analyzed to demonstrate the advantage of the proposed \textit{HybridGNN}.
\end{enumerate}

\section{Preliminary}
\label{sect:pre}

\begin{table}[t]
\scriptsize
\centering
\caption{Notations in this paper.}

\begin{tabular}{c|l}

\toprule
  Notations & Definitions \\
\midrule
    $\mathcal{V}$ & Node set in a graph \\
    $\mathcal{E}$ & Edge set in a graph, $\mathcal{E}\subset \mathcal{V}\times\mathcal{V}$\\
 $\mathcal{G}$ & A multiplex heterogeneous graph  $\mathcal{G}=(\mathcal{V}, \mathcal{E},\phi,\psi)$ \\
 $\mathcal{O},\mathcal{R}$ & Node type set and edge type (relationship) set in $\mathcal{G}$ \\  
 $g_r$ & The subgraph of $\mathcal{G}$ with edge type $r$ \\
 $\mathcal{P}$ & A metapath scheme \\
 $p$ & A metapath instance \\
 $\rho(v_i)$ & The metapath schemes that start from $v_i$ \\
 $\phi(\cdot), \psi(\cdot)$ & The mapping function of nodes and edges, respectively\\
 $\mathcal{N}_\mathcal{P}^{k}(v_i)$ & The $k$-th step metapath-guided neighbors of node $v_i$ w.r.t. $\mathcal{P}$ \\
 $h_{v_i | \mathcal{P}_z}^{(k)}$ & The embedding of $v_i$ under  $\mathcal{P}_z$ at $k$-th step sampling\\
 $AGG(\cdot, \cdot)$ & The aggregation function $AGG(h_{v_i}, \{h_{v_j}, v_j\in\mathcal{N}(v_i)\})$\\
\bottomrule
\end{tabular}

\label{tab:notations}
\end{table}

\begin{MyDef}[Heterogeneous Network]
    A heterogeneous network is defined as a network $\mathcal{G}=(\mathcal{V}, \mathcal{E})$ associated with a node type mapping function $\phi\colon \mathcal{V}\rightarrow \mathcal{O}$ and an edge type mapping function $\psi\colon \mathcal{E}\rightarrow \mathcal{R}$, and satisfies $\colon|\mathcal{O}|+|\mathcal{R}|>2$, where $\mathcal{O}$ and $\mathcal{R}$ represent the set of all node types and all edge types in $\mathcal{G}$, respectively. For convenience, the node set including all nodes with the same node type $\phi(v_i)$ as $v_i$ is denoted as $\kappa(v_i)=\{v_j|\phi(v_j)=\phi(v_i), \forall v_j\in \mathcal{V}\}$.
\end{MyDef}
\begin{MyDef}[Multiplex Heterogeneous Network]
    A multiplex heterogeneous network is defined as a heterogeneous network $\mathcal{G}=(\mathcal{V}, \mathcal{E})$, $\mathcal{E} \subseteq \mathcal{V}\times\mathcal{V}\times\mathcal{R}$, where $|\mathcal{R}|>1$.
    Different from traditional heterogeneous networks, there can be multiple relationships (\ie, edges) between two nodes in a multiplex heterogeneous network.
    \label{def:han}
\end{MyDef}

For example, as shown in Fig.~\ref{fig:intro_example}, $u_1$ and $a_1$ are connected not only under the \textit{like} relationship, but also under the \textit{comment} relationship.

The diversity of interactions among nodes requires relationship-specific representation learning methods to have the ability of capturing user preferences under various relationships. 
We also denote the relationship-specific subgraph of $\mathcal{G}$ as $g_{r}, r\in\mathcal{R}$, where no other relationships except $r$ exist.

With the above definitions, the goal of multiplex heterogeneous network embedding is to learn a low-dimensional representation $e^*_{v_i, r} \in \mathbb{R}^{d}$ ($d \ll |\mathcal{V}|$) for each node $v_i$ under each relationship $r \in \mathcal{R}$.

\begin{MyDef}[Metapath Scheme]
    A metapath scheme $\mathcal{P}$ is defined as a path with specified node types and edge types, denoted as $\mathcal{P} = o_0\stackrel{r_1}{\rightarrow}o_1\stackrel{r_2}{\rightarrow} \cdots \stackrel{r_n}{\rightarrow}o_n$, where $o_i\in\mathcal{O}$ and $r_i\in \mathcal{R}$, $|\mathcal{P}|=n$. Specifically, if $r_1=r_2=\cdots=r_n$, we call $\mathcal{P}$ an intra-relationship metapath scheme, otherwise an inter-relationship metapath scheme.
    \label{def:meta_scheme}
\end{MyDef}

\begin{MyDef}[Metapath Instance]
    Given a metapath scheme $\mathcal{P}=o_0\stackrel{r_1}{\rightarrow}o_1\stackrel{r_2}{\rightarrow} \cdots \stackrel{r_n}{\rightarrow}o_n$, a metapath instance $p$ of $\mathcal{P}$ is defined as a sampled node sequence $p=v_0\stackrel{r_1}{\rightarrow}v_1\stackrel{r_2}{\rightarrow}\cdots \stackrel{r_n}{\rightarrow}v_n$, \todo{which satisfies the scheme $\mathcal{P}$, i.e., $\phi(v_0) = o_0, \phi(v_1) = o_1, \cdots, \phi(v_n) = o_n$}.
   Moreover, we call $p$ an intra-relationship metapath instance if it is sampled under an intra-relationship metapath scheme, otherwise we call $p$ an inter-relationship metapath instance.
    \label{def:meta_inst}
\end{MyDef}

In multiplex heterogeneous networks, metapath schemes and their corresponding metapath instances are important to explicitly reveal interactions among nodes. For example, a $User\stackrel{like}{\rightarrow}Author\stackrel{like}{\rightarrow}User$ metapath scheme under the \textit{like} relationship illustrates users' common preferences of favored authors.

\begin{MyDef}[Metapath-guided Neighbors]
    Given a node $v_i$ and a metapath scheme $\mathcal{P}$ in an HIN, a metapath-guided neighbor of $v_i$ is defined as one of all visited nodes when $v_i$ walks along all metapath instances that satisfy $\mathcal{P}$. 
    In addition, we denote all metapath-guided neighbors of $v_i$ after the $k$-th step as $\mathcal{N}_\mathcal{P}^{k}(v_i)$. Specifically, $\mathcal{N}_\mathcal{P}^{0}(v_i)=\{v_i\}$.
\end{MyDef}

Please note that different from similar concepts proposed in \cite{wang2019heterogeneous, dong2017metapath2vec}, metapath-guided neighbors aim to find a specific aggregation flow that follows a certain metapath scheme $\mathcal{P}$ with heterogeneity preserved, rather than find similar nodes with $v_i$ in HINs. In this case, metapath schemes used for metapath-guided neighbors can be asymmetric.
For simplicity, we use $\rho(v_i)$ to represent all metapath schemes starting from $v_i$.

As shown in Fig.~\ref{fig:intro_example}, take a metapath scheme $\mathcal{P} = Video\stackrel{like}{\rightarrow}User\stackrel{comment}{\rightarrow}Author$ as an example.
The metapath-guided neighbors of $v_1$ after the first, second and third step are $\mathcal{N}_\mathcal{P}^{0}(v_1)=\{v_1\}$, $\mathcal{N}_\mathcal{P}^{1}(v_1)=\{u_1, u_2\}$ and $\mathcal{N}_\mathcal{P}^{2}(v_1)=\{a_1\}$, respectively. 
Note that several important notations used in this paper are stated in Table \ref{tab:notations}.

\section{Methods}
\begin{figure*}[ht]
\centering
\includegraphics[width=1\linewidth]{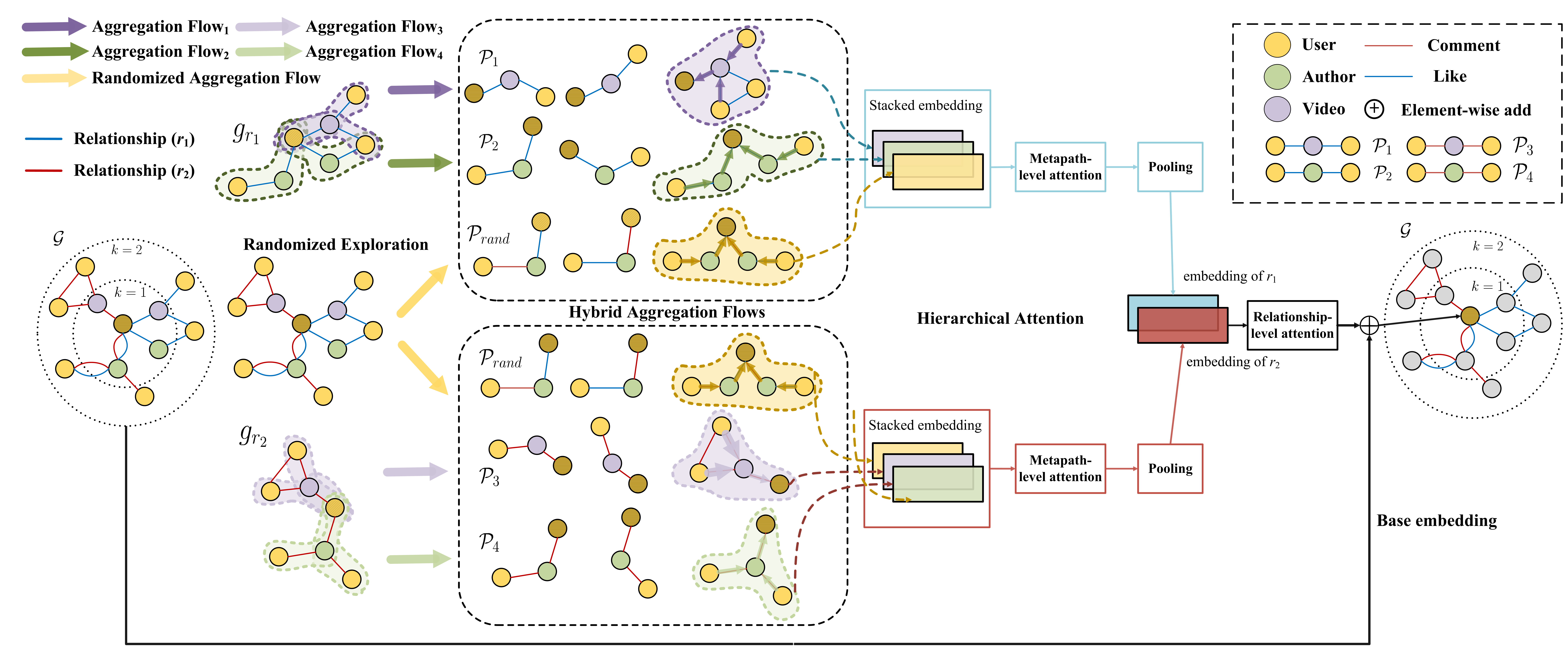} 
\caption{The framework of \textit{HybridGNN}. It first samples neighbors by each intra-relationship metapath and results from randomized exploration. Then, it performs hybrid aggregation flows to get the overall embeddings. We use hierarchical attentions (\ie, metapath-level attention and relationship-level attention) to distinguish the importance of heterogeneous messages and to learn latent connections among diverse relationships, respectively. Final representations are the concatenation with the base embedding.}
\label{fig:arch}
\end{figure*}
This section firstly presents the overview of our \textit{HybridGNN} model. 
Next, we formulate randomized inter-relationship exploration, hybrid aggregation flows and the hierarchical attention module. 
Then, the model training is discussed. 
After that, we analyze the attention mechanism and the time complexity of \textit{HybridGNN}.
Finally, we discuss several situations of \textit{HybridGNN} and compare \textit{HybridGNN} with three related works.

\subsection{Overview of \textit{HybridGNN}}
\label{sect:overview}

We first present the framework of \textit{HybridGNN}. As shown in Fig.~\ref{fig:arch}, the \textit{HybridGNN} model mainly consists of three parts: randomized inter-relationship exploration, hybrid aggregation flows and hierarchical attention modules.

The randomized inter-relationship exploration module adopts a two-phase sampling method to gather inter-relationship neighbors. The results of the module are used as auxiliary information to explore latent interactive patterns in $\mathcal{G}$.

In the hybrid aggregation flows module, we integrate information of nodes according to predefined metapaths as well as the results from randomized inter-relationship exploration. 
Thus, each subgraph $g_r$ has multiple aggregation flows. 
Specifically, each aggregation flow represents specific semantic information from both predefined metapaths that are emphasized to \todo{be captured} in $\mathcal{G}$ and the results of randomized inter-relationship exploration that discovers latent patterns in $\mathcal{G}$. 

The hierarchical attention module consists of metapath-level attention and relationship-level attention. Metapath-level attention applies attention mechanisms to embeddings from hybrid aggregation flows. 
It is used to distinguish the importance of heterogeneous message passing from that of metapath-based and randomized exploration.
Relationship-level attention applies attention mechanisms on node embeddings under every relationship to learn latent connections among diverse relationships.

\textit{HybridGNN} finally outputs the node embedding $e^*_{v_i, r_l}$ under a specific relationship $r_l$ for downstream tasks. Note that in this paper, we use the embedding $e^*_{v_i, r_l}$ for the link prediction task in multiplex heterogeneous network $\mathcal{G}$.




\subsection{Randomized Inter-relationship Exploration}
\label{sect:random}
This subsection proposes the randomized inter-relationship exploration module which adopts a two-phase sampling strategy. 
Briefly, given \todo{a starting point $v_0$}, the randomized inter-relationship exploration module first samples a relationship (\todo{say} $r'$) on $\mathcal{G}$, and then it samples nodes (\todo{say} \{$u^1$, $u^2$, \dots \}) which are endpoints of edges with the relationship $r'$.
Let $v_t$ be the $t$-th node in a walk from $v_0$ to one of the sampled nodes (\todo{say} $u^1$), \ie, $v_0\stackrel{r_1}{\rightarrow}v_1\stackrel{r_2}{\rightarrow}\cdots \stackrel{r_t}{\rightarrow}v_t\stackrel{r_{t+1}}{\rightarrow} v_{t+1}\stackrel{r_{t+2}}{\rightarrow}\cdots\stackrel{r'}{\rightarrow}u^1$.
Then, the transition between $v_t$ and $v_{t+1}$ is regarded as a joint probability, \ie, $ p(r_{t+1}|v_t)\times p(v_{t+1}|r_{t+1}, v_t)$.

Specifically, $r_{t+1}$ is chosen by the following distribution: 
\begin{equation}
    p(r_{t+1}|v_t)=
    \begin{cases}
    \frac{1}{|\{r|N_{r}(v_{t})\neq\emptyset,\forall r\in\mathcal{R}\}|} & N_{r_{t+1}}(v_{t})\neq\emptyset \\
    0 & N_{r_{t+1}}(v_{t})=\emptyset ,
    \end{cases}
\end{equation}
where $N_r(v_t)$ denotes the neighbors of $v_t$ under relationship $r$.

Then, $v_{t+1}$ is sampled as follows: 
\begin{equation}
p(v_{t+1}|r_{t+1}, v_t)=\frac{1}{|N_{r_{t+1}}(v_{t})|} \;,
\end{equation}
where $N_{r_{t+1}}(v_t)$ denotes the neighbors of $v_t$ under relationship $r_{t+1}$.

\subsection{Hybrid Aggregation Flows}
\label{sect:intra}

We denote the predefined metapath scheme set under relationship $r_l$ as $PS_{r_l}=$ $\{\mathcal{P}_1, \mathcal{P}_2,$ $\cdots, \mathcal{P}_n\}$, $|PS_{r_l}|=n$.
Then the metapath scheme set that is satisfied by metapath instances starting from $v_i$ is $\rho(v_i) \cap PS_{r_l}$.
Thus for each node $v_i$ in $g_{r_l}$ and each metapath scheme $\mathcal{P}_z\in \rho(v_i) \cap PS_{r_l}$, we have the metapath-guided neighbors $\mathcal{N}_{\mathcal{P}_z}^{k}(v_i)$.
As $PS_{r_l}$ preserves explicit heterogeneous semantics, we adopt a metapath scheme specified aggregation to learn the edge embedding of $v_i$ for each metapath scheme $\mathcal{P}_z\in \rho(v_i) \cap PS_{r_l}$, $|\mathcal{P}_z|=K$, denoted as $h_{v_i | \mathcal{P}_z}^{(K)}$. It can be recursively formulated as:
\begin{small}
\begin{equation}
\label{eq:intra_agg}
     h_{v_i | \mathcal{P}_z}^{(k)} = AGG_{\mathcal{P}_z}(h_{v_i| \mathcal{P}_z}^{(k-1)}, \{h_{v_j|\mathcal{P}_z}^{(k-1)}\colon v_j\in \mathcal{N}_{\mathcal{P}_z}^{K-k+1}(v_i)\}),
\end{equation}
\end{small}where $h_{v_j| \mathcal{P}_z}^{(0)}, v_j\in \mathcal{N}_{\mathcal{P}_z}^{K}(v_i)$ is randomly initialized, $k\in [1,2,\cdots, K]$, $K$ is the maximum step of sampling starting from $v_i$ and $h_{v_i| \mathcal{P}_z}^{(k-1)}$ is the $d_h$-dimensional edge embedding of node $v_i$ under metapath scheme $\mathcal{P}_z$ at $(k-1)$-th step sampling. \textit{AGG} is the aggregation function which aggregates information from the $k$-th metapath neighbors. There are normally three aggregation candidates, \ie, mean aggregator, LSTM aggregator and pooling aggregator. In the rest of the paper, we focus on the mean aggregator as there are no significant differences among these aggregators in our experiments.
\label{sect:inter}

We also apply the aggregation function to aggregate information from the results by the randomized inter-relationship exploration starting from $v_i$. Therefore, $h_{v_i | \mathcal{P}_{rand}}^{(k)}$ is calculated as:
\begin{small}
\begin{equation}
\label{eq:inter_agg}
\begin{aligned}
     AGG_{\mathcal{P}_{rand}}\left(h_{v_i| \mathcal{P}_{rand}}^{(k-1)}, \left\{h_{v_j| \mathcal{P}_{rand}}^{(k-1)}\colon v_j\in \mathcal{N}_{\mathcal{P}_{rand}}^{K-k+1}(v_i) \right\}\right),
\end{aligned}
\end{equation}
\end{small}where $h_{v_j | \mathcal{P}_{rand}}^{(0)}, v_j\in\mathcal{N}_{\mathcal{P}_{rand}}^{K}(v_i)$ is also randomly initialized, $k\in[1,2,\cdots, K]$ and $K$ is the maximum step of sampling starting from $v_i$, $h_{v_i | \mathcal{P}_{rand}}^{(k)}$ is a $d_h$-dimensional edge embedding of node $v_i$. Different from $\mathcal{P}_{{r_l}}$, path instances in $\mathcal{P}_{rand}$ do not follow a specified metapath scheme and learnable weights are shared among the randomized sample neighbors.


Then, hybrid aggregation module concatenates these edge embeddings of $v_i$ under $\rho(v_i) \cap PS_{r_l}$ (i.e., $h_{v_i | \mathcal{P}_z}^{(K)}$) and the results of the randomized inter-relationship exploration (i.e., $h_{v_i | \mathcal{P}_{rand}}^{(k)}$) to gather comprehensive information related to $v_i$.
\begin{small}
\begin{equation}
\label{eq:before_hpv}
    H_{\rho(v_i)} = CONCAT\left(\{h_{v_i | \mathcal{P}_{rand}}^{(k)}, h_{v_i | \mathcal{P}_j}^{(k)}\colon \mathcal{P}_j \in \rho(v_i)\}\right),
\end{equation}
\end{small}where $H_{\rho(v_i)}\in \mathbb{R}^{(|\rho\left(v_i)|+1\right)\times d_h}$.

\subsection{Hierarchical Attention}
\label{sect:hie}
In the metapath-level attention, we adopt self-attention mechanism \cite{vaswani2017attention} to learn the weights among metapath schemes and randomized inter-relationship walks.
\begin{tiny}
\begin{equation}
\label{eq:after_hpv}
\begin{aligned}
     \hat{H}_{\rho(v_i)}  
     &=\mathcal{A}(Q_{\rho(v_i)},K_{\rho(v_i)},V_{\rho(v_i)}) \\
     &=softmax(\frac{H_{\rho(v_i)}W_{Q_{\rho(v_i)}}\cdot H_{\rho(v_i)}W_{K_{\rho(v_i)}}}{\sqrt{d_k}})H_{\rho(v_i)}W_{V_{\rho(v_i)}},
\end{aligned}
\end{equation}
\end{tiny}where $Q_{\rho(v_i)}=H_{\rho(v_i)}$, $K_{\rho(v_i)}=H_{\rho(v_i)}$, $V_{\rho(v_i)}=H_{\rho(v_i)}$, $W_{Q_{\rho(v_i)}}$, $W_{K_{\rho(v_i)}}$, $W_{V_{\rho(v_i)}}$ $\in\mathbb{R}^{d_h\times d_k}$ are the trainable parameters, $d_k$ is the hidden embedding size and $\hat{H}_{\rho(v_i)}\in\mathbb{R}^{(|\rho\left(v_i)|+1\right)\times d_k}$.

Note that $\hat{H}_{\rho(v_i)}$ can be regarded as a concatenation of re-weighted embedding $\hat{h}_{v_i | \mathcal{P}_{rand}}^{(k)}$ and $\hat{h}_{v_i | \mathcal{P}_j}^{(k)}$, $\mathcal{P}_j \in \rho(v_i) \cap PS_{r_l} $ after self-attention along the metapath dimension (\ie, $|\rho(v_i)|+1$). Therefore, the embedding of node $v_i$ under a specific relationship $r_l$ is defined as the mean pooling of $\hat{h}_{\rho(v_i)}$:
\begin{small}
\begin{equation}
\label{eq:embed_rela}
     \hat{h}_{v_i, {r_l}} = \frac{1}{|\rho(v_i)|+1}\left( \hat{h}_{v_i | \mathcal{P}_{rand}}^{(k)} + \sum_{\mathcal{P}_j \in \rho(v_i)\todo{\cap PS_{r_l}}} \hat{h}_{v_i | \mathcal{P}_j}^{(k)}\right).
\end{equation}
\end{small}

After metapath-level attention applied on the embeddings by hybrid aggregation flows, inter-relationships are fused into $\hat{h}_{v_i, {r_l}}$. We then concatenate these embeddings under each relationship.
\begin{equation}
\label{eq:before_ur}
    U_{v_i, \mathcal{R}} = CONCAT\left(\{\hat{h}_{v_i, {r_l}}\colon r_l \in \mathcal{R}\}\right).
\end{equation}

In the relationship-level attention, we also apply the self-attention mechanism to learn the weights among relationships.
\begin{tiny}
\begin{equation}
\label{eq:after_ur}
\begin{aligned}
     \hat{U}_{v_i, \mathcal{R}} 
     &=\mathcal{A}(Q_{v_i, \mathcal{R}},K_{v_i, \mathcal{R}},V_{v_i, \mathcal{R}}) \\
     &=softmax\left(\frac{U_{v_i, \mathcal{R}}W_{Q_{v_i, \mathcal{R}}}\cdot U_{v_i, \mathcal{R}}W_{K_{v_i, \mathcal{R}}}}{\sqrt{d_k}}\right)U_{v_i, \mathcal{R}}W_{V_{v_i, \mathcal{R}}},
\end{aligned}
\end{equation}
\end{tiny}where $Q_{v_i, \mathcal{R}}=U_{v_i, \mathcal{R}}$, 
$K_{v_i, \mathcal{R}}=U_{v_i, \mathcal{R}}$,
$V_{v_i, \mathcal{R}}=U_{v_i, \mathcal{R}}$,
$W_{Q_{v_i, \mathcal{R}}}$, 
$W_{K_{v_i, \mathcal{R}}}$, 
$W_{V_{v_i, \mathcal{R}}}$ $\in\mathbb{R}^{d_k\times d_k}$ are the trainable parameters, $d_k$ is the hidden embedding size and $\hat{U}_{v_i, \mathcal{R}}\in\mathbb{R}^{|\mathcal{R}|\times d_k}$.

Please note that $\hat{U}_{v_i, \mathcal{R}}$ contains all the relationship-specific local embeddings of node $v_i$, \ie, $\hat{U}_{v_i, \mathcal{R}} =\left(e_{v_i, r_1}, e_{v_i, r_2}, \cdots, e_{v_i, r_L}\right), |\mathcal{R}|=L$. 
Then, the final embedding of node $v_i$ under a specific relationship $r_l$ is written as:
\begin{equation}
    e^*_{v_i, r_l} = e_{v_i} + e_{v_i, r_l}\cdot W_{v_i,r_l},
\end{equation}where $e_{v_i}$ is the basic embedding, $e_{v_i, r_l}$ is the local edge embedding for node $v_i$ under relationship $r_l$, and $W_{v_i, r_l}\in\mathbb{R}^{d_k\times d}$ is the weight matrix to be learned. 

\subsection{Model Training}
We discuss the learning procedure of \textit{HybridGNN}. Following \cite{dong2017metapath2vec, grover2016node2vec}, in order to better measure the similarity between nodes in multiplex heterogeneous networks, metapath-based random walks are adopted. According to the transition probability at position $t$ of the path, we define the following metapath scheme for every relationship $r$ as
$\mathcal{P}=\phi(v_0)\stackrel{r}{\rightarrow}\phi(v_1)\stackrel{r}{\rightarrow}\phi(v_2)\cdots \stackrel{r}{\rightarrow}\phi(v_n)$, where $n$ is the length of the metapath.
%
\begin{tiny}
\begin{equation}
\begin{aligned}
    \mathcal{T}(v_{t+1}|v_{t}) =
    \begin{cases}
    \frac{1}{|N_r(v_{t}) \cap\kappa(v_{t+1}) |} & (v_{t}, v_{t+1}) \in \mathcal{E}, \phi(v_{t+1})\in\kappa(v_{t+1}) \\
    0 &  (v_{t}, v_{t+1}) \in \mathcal{E}, \phi(v_{t+1})\notin \kappa({v}_{t+1}) \\
    0 &  (v_{t}, v_{t+1}) \notin \mathcal{E},
    \end{cases}
\end{aligned}
\end{equation}
\end{tiny}where $v_{t}, v_{t+1}\in \mathcal{V}$,
$N_r(v_t)$ denotes the neighbors of node $v_t$ on relationship $r$,
and $\kappa(v_{t+1})$ denotes the node set with same node type as $\phi(v_{t+1})$. Metapath based sampling enables to capture the semantic latent representation among different node types along the path with the help of heterogeneous skip-gram model.
The context $C(v_i)$ of a node $v_i$ is denoted as $C(v_i)=\{v_k|v_k\in \mathcal{S}, |k-i|\le \delta, k\neq i\}$, where $\delta$ is a threshold representing the radius of the moving window size,
and \todo{$\mathcal{S}$ is the node set along the sampled paths}. 
Please note that these paths are also used in the aggregation phases.

Our goal is to maximize the occurrence of a node $v_i$ with its context, and thus the objective function is to minimize the following negative log-likelihood:
\begin{equation}
\label{eq:object_func}
    -\log p_\theta(\prod_{v_k\in C} v_k|v_i) = -\sum_{v_k\in C}\log p_\theta (v_k|v_i),
\end{equation}
where $\theta$ denotes the parameters in the model.

\begin{algorithm}[t]
\caption{The Forward propagation of \textit{HybridGNN}.}
\label{alg:forward_prop}
\LinesNumbered 
\KwIn{
Multiplex Heterogeneous Graph $\mathcal{G}=(\mathcal{V}, \mathcal{E})$ ; 
Training batch set $\mathcal{V}_T$; 
Relationships $\mathcal{R}$ ;
Predefined metapath scheme set $\mathcal{P}_s=\{\mathcal{P}_1, \mathcal{P}_2, \cdots, \mathcal{P}_m\}$, where $m$ is the user-specified constants ; 
Metapath based aggregation layer $K_{z}$ ; 
Randomized inter-relationship exploration aggregation layer $K_{{rand}}$ ; 
Walk length $S$ ; 
Number of walks $N$ ; 
}
\KwOut{The embedding of each node $v_i \in \mathcal{V}$ under each relationship $r_l \in \mathcal{R}$, \ie, $e_{v_i, r_l}^*$}
$e_{v_i}\leftarrow W x_v$, $\forall v_i\in \mathcal{V}$ \; 
\For{$r_l$, $r_l\in\mathcal{R}$}{
\For{$\mathcal{P}_z$,
$\mathcal{P}_z\in\mathcal{P}_s$}{
    \For{$k=1\cdots K_{z}$}{
        \For{$v_i\in\mathcal{V}_T$}{
            $h_{v_i | \mathcal{P}_z}^{(k)} = AGG_{\mathcal{P}_z}(h_{v_i| \mathcal{P}_z}^{(k-1)}, \{h_{v_j|\mathcal{P}_z}^{(k-1)}\colon v_j\in \mathcal{N}_{\mathcal{P}_z}^{K_{z}-k+1}(v_i) \})$ \;
        }
    }
}
\For{$k=1\cdots K_{{rand}}$}{
    \For{$v_i\in\mathcal{V}_T$}{
        $h_{v_i | \mathcal{P}_{rand}}^{(k)} = AGG_{\mathcal{P}_{rand}}(h_{v_i| \mathcal{P}_{rand}}^{(k-1)}, \{h_{v_j| \mathcal{P}_{rand}}^{(k-1)}\colon v_j\in \mathcal{N}_{\mathcal{P}_{rand}}^{K_{{rand}}-k+1}(v_i) \})$\;
    }
}
    Concatenating the embedding of  $h_{v_i | \mathcal{P}_{rand}}^{(K_{\mathcal{P}_{rand}})}$, and $h_{v_i | \mathcal{P}_z}^{(K_{\mathcal{P}_z})}$, $\mathcal{P}_z\in\rho(v_i)$, according to Eq. \ref{eq:before_hpv}\;
    Applying the self-attention mechanism to the representation of $v_i$ under all the latent metapath schemes, \ie, $H_{\rho(v_i)}$, according to Eq. \ref{eq:after_hpv} \;
    Learning the embedding of $v_i$ under $r_l$, according to Eq. \ref{eq:embed_rela}.
}
Concatenating the embedding of $\hat{h}_{v_i, {r_l}}$, $r_l\in\mathcal{R}$, according to Eq. \ref{eq:before_ur}\;
Applying the self-attention mechanism to the representation of $v_i$ under all relationships, \ie, $U_{v_i,\mathcal{R}}$, according to Eq. \ref{eq:after_ur}. \;
$e^*_{v_i, r_l} = e_{v_i} + e_{v_i, r_l}\cdot W_{v_i, r_l}$\;
\Return{$e^*_{v_i, r_l}, \forall v_i \in \mathcal{V}, \forall r_l \in \mathcal{R}$}
\end{algorithm}

For computational efficiency, we follow metapath2vec~\cite{dong2017metapath2vec} and adopt heterogeneous negative sampling to approximate Equation \ref{eq:object_func}. Therefore, the final objective function is defined as:
\begin{scriptsize}
\begin{equation}
        \mathcal{L}=-\log \sigma(c_j\cdot e^*_{v_i, r_l}) - \sum_{l=1}^{L}\mathbb{E}_{v_{k}\sim P_{Neg}}\left[\log \sigma(-c_k\cdot e^*_{v_i, r_l}) \right],
\end{equation}
\end{scriptsize}where $c_j$ is the context embedding of node $j$, $\sigma(\cdot)$ is the sigmoid function, i.e., $\sigma(x)=\frac{1}{1+e^{-x}}$, $P_{Neg}$ is a noise distribution in negative sampling and $v_k$ is the node sampled from $P_{Neg}$. The pseudo code of forward propagation of \textit{HybridGNN} is presented in Algorithm \ref{alg:forward_prop}.

\subsection{Model Analysis}

We first analyze the mechanism of applying self-attention to learn comprehensive and attentive information from all schemes $\rho(v_i)$ of $v_i$. Consider the situation where no inter-relationship information exists in the multiplex heterogeneous network $\mathcal{G}$. In this case, \textit{HybridGNN} only learns semantics from intra-relationship schemes during aggregation. According to Equation \ref{eq:before_hpv} and Equation \ref{eq:after_hpv}, $\hat{H}_{\rho(v_i)} $ can be written as:
\begin{equation}
    \hat{H}_{\rho(v_i)} = CONCAT\left(\left\{\alpha_j\cdot h_{v_i | \mathcal{P}_j}^{(k)}\colon \mathcal{P}_j \in \rho(v_i)\right\}\right),
\end{equation}
where $\alpha_j$ is the attention weight of the metapath scheme $\mathcal{P}_j$ and $\sum \alpha_j=1$. 

Therefore, the multiplexity property of $\mathcal{G}$ is not utilized in the metapath-level. It is only utilized in the relationship-level interaction, \ie, in Equation \ref{eq:before_ur} and Equation \ref{eq:after_ur}.
However, it is considered in \textit{HybridGNN} by the introduction of the randomized inter-relationship exploration and the metapath-level attention. Specifically, $\hat{H}_{\rho(v_i)}$ is calculated as:
\begin{scriptsize}
\begin{equation}
\begin{aligned}
    CONCAT\left(\{\beta_{v_i} h_{v_i | \mathcal{P}_{rand}}^{(k)}, (1-\beta_{v_i}) h_{v_i | \mathcal{P}_j}^{(k)}\colon \mathcal{P}_j \in \rho(v_i)\}\right),
\end{aligned}
\end{equation}
\end{scriptsize}where $1-\beta_{v_i}=\sum \alpha_j$, $\beta_{v_i}$ is a scalar in the range $[0, 1]$ and it is regarded as a randomized exploration factor that controls the influence of inter-relationship information during aggregation.

We also analyze the time complexity of \textit{HybridGNN}. Overall, \textit{HybridGNN} has two time-consuming parts, \ie, hybrid aggregation flows and hierarchical attention. Considering a multiplex heterogeneous network $\mathcal{G}=(\mathcal{V}, \mathcal{E})$ with $|\mathcal{O}|$ node types and $|\mathcal{R}|$ edge types, the maximal aggregation layer is $K_{agg}$ and we sample fixed number of neighbors $N_k$ at each layer $k$. Then the time complexity during hybrid aggregation is $\prod_{i=1}^{K_{agg}} N_i\cdot d_k^2$. 
As the hierarchical attention consists of metapath-level attention and relationship-level attention, the time complexity is $O(|\rho(v)+1|^2\cdot d_k)+O(|\mathcal{R}|^2\cdot d_k)$. Therefore, the time complexity of \textit{HybridGNN} is $\prod_{i=1}^{K_{agg}} N_i\cdot d_k^2 + O(|\rho(v)+1|^2\cdot d_k)+O(|\mathcal{R}|^2\cdot d_k)$.

\normalsize

\subsection{Categorization and Comparison}

In this subsection, we discuss several situations of the proposed \textit{HybridGNN} in the heterogeneous network $\mathcal{G}$. Then, we choose three related works used in HINs to show the differences between \textit{HybridGNN} and previous studies.

\vspace{4pt}
\noindent{\textbf{Categorization of Heterogeneous Network.}}
We discuss three types of heterogeneous networks, which cover most of the situations where graph neural networks are applied. 

a) In heterogeneous network $\mathcal{G}_1=(\mathcal{V}, \mathcal{E})$, where $|\mathcal{O}|=1$ and $|\mathcal{R}|\ge 2$.
Therefore, after $\mathcal{G}_1$ is split by different relationships, $\mathcal{G}_1$ is induced into several relationship-specific \todo{subgraphs} $g_i$. The total number of $g_i$ is $|\mathcal{R}|$. As $\mathcal{G}_1$ only has one node type, pre-defined metapaths degrade to a path $p$ where each node type in $p$ is the same. In this situation, the performance of metapaths based aggregation approximates to the performance based on random walk.
We call $\mathcal{G}_1$ the graph with heterogeneous information on relationships but with homogeneous information on nodes.

b) In heterogeneous network $\mathcal{G}_2=(\mathcal{V}, \mathcal{E})$, where $|\mathcal{O}|\ge 2$ and $|\mathcal{R}| = 1$.
In this situation, we find that the \textit{hierarchical attention} module takes little effect as the whole graph $\mathcal{G}_2$ cannot be induced to 
subgraphs according to its type (\ie, $|\mathcal{R}|=1$). However, we can discover many meaningful schemes on such graph as $|\mathcal{O}|\ge 2$. For example, we can define two paths like $U\stackrel{r}{\rightarrow}V\stackrel{r}{\rightarrow}U$,  $U\stackrel{r}{\rightarrow}A\stackrel{r}{\rightarrow}U$, \todo{where $U$, $V$ and $A$ denote the node type of \textit{User}, \textit{Video} and \textit{Author}}, respectively, and the relationship $r$ defined on these metapaths is \textit{View}. 
In this situation, as different metapaths reflect different user behaviors and the corresponding preferences, our \textit{hybrid aggregation flows} module will learn such information along two different metapaths with a self-attention mechanism to balance the weights. We call such graph $\mathcal{G}_2$ the graph with heterogeneous information on nodes but with homogeneous information on relationships.

c) In heterogeneous network $\mathcal{G}_3=(\mathcal{V}, \mathcal{E})$, where $|\mathcal{O}|\ge 2$ and $|\mathcal{R}| \ge 2$.
It is a more general case in heterogeneous networks compared to the above two situations.
In this situation, our proposed modules (\ie, \textit{randomized inter-relationship exploration}, \textit{hybrid aggregation flows} and \textit{hierarchical attention}) take effect and achieve an optimal performance. 
Specifically, \textit{randomized inter-relationship exploration module} is to handle the heterogeneity of edges, and \textit{hybrid aggregation flows module} is to handle the heterogeneity of nodes. 
These approaches contribute to the node representation learning in $\mathcal{G}_3$. We call such graph $\mathcal{G}_3$ the graph with heterogeneous information on both nodes and relationships.

Please also note a special case $\mathcal{G}_4=(\mathcal{V}, \mathcal{E})$, where $|\mathcal{O}| = 2$ and $\mathcal{G}_4$ is a bipartite graph.
In this condition, the metapath guided neighbors sampled at $k$-th step started with the same node type have the same node type due to the bipartite assumptions.
$\mathcal{G}_4$ makes the metapaths based aggregation (\ie, \textit{hybrid aggregation flows}) degrade into random walk based aggregation unless different orders of paths 
are adopted.

\vspace{4pt}
\noindent{\textbf{Comparison with Previous Works.}}
We choose three related works used in heterogeneous networks to show the differences between our model and previous studies. The three models are RGCN\cite{schlichtkrull2018modeling}, GATNE\cite{cen2019representation} and MAGNN\cite{fu2020magnn}.

\todo{\textbf{RGCN}. RGCN is a statistical representation learning (SRL) model to predict missing information in knowledge base. 
It learns node embeddings through relational layer aggregation (\ie, $h_i^{(l+1)}=\sigma(\sum_{r\in\mathcal{R}}\sum_{j\in N_i^r}\frac{1}{c_{i, r}}W_r^{(l)}h_j^{(l)}+W_0^{(l)}h_i^{(l)})$, where $N_i^r$ represents the neighbors of node $i$ under edge type $r$). However, this aggregation approach is equivalent to adopting random neighbor sampling under each relationship-specific subgraph. Thus it only treats the complex interactions of nodes in a node-level manner while it neglects to explicitly learn the useful metapath in a path-level manner.} 
 
\textbf{GATNE}. GATNE considers that node embeddings in heterogeneous networks are relationship-specific, which can be simplified as $m_{i, r_i}=b_i+\alpha W m'_{i, r_i}$, where $b_i$ is a shared base embedding and $m'_{i, r_i}$ is the relationship-specific embedding for a certain relationship $r_i$. However, \todo{as this work ignores the heterogeneity of the nodes, it cannot fully capture the heterogeneity of the interaction among nodes.}
\todo{Moreover, the ignorance of nodes' heterogeneity further fails to model the diversity of metapaths.}


We show that when a heterogeneous network $\mathcal{G}$ has no more than two node types (\ie, $|\mathcal{O}|\le 2$), the metapaths are trivial and can be omitted. For example, given metapath schemes like $\mathcal{P}_1=U\rightarrow I\rightarrow U$ \todo{when $\mathcal{G}$ is a bipartite graph} or $\mathcal{P}_2=U\rightarrow U$ (\ie, $|\mathcal{O}|=1$), the effect of \textit{hybrid aggregation flows} is degraded to the vanilla neighbor sampling method in some cases. 
\todo{As the node type is known at each sampling step of metapath-guided neighbors, the diversity of metapaths only has to rely on the different orders of paths}.

In this aspect, our model is a generalized model of GATNE when the heterogeneous graph $\mathcal{G}$ has more than two node types, (\ie, $|\mathcal{O}|>2$). 
\todo{Furthermore, compared to GATNE, the introduction of \textit{randomized exploration} can prevent overfitting and improve the generalization ability of \textit{HybridGNN} when $\mathcal{G}$ is a bipartite graph or has only one node type}.
We make more discussions on the heterogeneity of $\mathcal{G}$ in terms of the number of relationships $|\mathcal{R}|$ and node types $|\mathcal{O}|$ in the experiment. 

\begin{table}[t]
\centering
\caption{The statistics of datasets used in our experiments.}

\begin{tabular}{c|c|c|c|c|c}
\toprule
Datasets  & $|\mathcal{V}|$ & $|\mathcal{E}|$ & $|\mathcal{O}|$ & $|\mathcal{R}|$ & $\mathcal{P}$ \\
\midrule
Amazon & 10,099 & 148,659 & 1 & 2 & I-I-I \\
\midrule
YouTube & 2,000 & 1,310,544 & 1 & 5 & I-I-I  \\
\midrule
\multirow{6}{*}{IMDb} & \multirow{6}{*}{11,616} & \multirow{6}{*}{34,212} & \multirow{6}{*}{3} & \multirow{6}{*}{1} & M-D-M \\
& & & & & M-A-M \\
& & & & & D-M-D \\
& & & & & A-M-A \\
& & & & & D-M-A-M-D\\
& & & & & A-M-D-M-A\\
\midrule
\multirow{2}{*}{Taobao} & \multirow{2}{*}{64,737} & \multirow{2}{*}{144,511} &
\multirow{2}{*}{2} & \multirow{2}{*}{4} & U-I-U \\
& & & & & I-U-I \\
\midrule
\multirow{4}{*}{Kuaishou} & \multirow{4}{*}{105,749} & \multirow{4}{*}{175,870} & \multirow{4}{*}{3} & \multirow{4}{*}{4} & U-A-U \\
& & & & & A-U-A \\ 
& & & & & V-U-V \\
& & & & & U-V-U \\
\bottomrule
\end{tabular}
\label{tab:datasets}
\end{table}

\textbf{MAGNN}. MAGNN is also proposed to tackle the problem of heterogeneous node embedding in view of the metapath based aggregation.
Considering the transductive form of MAGNN, two main aggregation operations in MAGNN are the intra-metapath aggregation and inter-metapath aggregation, where the former aggregates information along a metapath and the latter aggregates information among metapaths.

However, the node embedding in MAGNN is a strict metapath-based aggregation approach, which over-emphasizes the importance of a sampled path rather than the inherent local structure of the heterogeneous $G$ itself.
Specifically, MAGNN learns the embedding  of a metapath through $h_p=f_\theta(p)$, where $f_\theta$ is a learnable function that outputs the embedding of a metapath instance. \todo{Our model, on the contrary, assumes that the $k$-th metapath-guided neighbors sampled from a same metapath scheme should share one aggregation flow, (\ie, $h_{v_i | \mathcal{P}_{rand}}^{(k)}=AGG_{\mathcal{P}_{rand}}\left(h_{v_i| \mathcal{P}_{rand}}^{(k-1)}, \left\{h_{v_j| \mathcal{P}_{rand}}^{(k-1)}\colon v_j\in \mathcal{N}_{\mathcal{P}_{rand}}^{K-k+1}(v_i) \right\}\right)$, where $AGG(\cdot, \{\cdots\})$ is the aggregation function). This operation can be viewed as a local parameter sharing among these metapath instances.
For inductive learning, \textit{HybridGNN} can leverage the advantages between node features and the topological structure of node neighbors}. It indicates our model is intended to learn the embedding more from its metapath-guided neighbors rather than the aggregation of sampled metapath instances. 


\section{Experiments}
\label{sect:exp}
In this section, the statistics of the datasets, baselines as well as the experimental settings are firstly introduced. Then the comparison between \textit{HybridGNN} and other baselines on the link prediction task is conducted, and the performances are carefully analyzed. 
Moreover, the effect of the randomized exploration and the uplift from inter-relationship are analyzed. The ablation study is also conducted to confirm the effect of each module in \textit{HybridGNN}.
Finally, parameter sensitivity and case study are discussed.

\subsection{Datasets}

We evaluate the performance of \textit{HybridGNN} on five real-world datasets, including four public datasets (\ie, Amazon \cite{cen2019representation, he2016ups,mcauley2015image}, YouTube \cite{tang2009uncovering}, IMDB \cite{fu2020magnn}, Taobao \cite{zhu2018learning, zhu2019joint}) and one Kuaishou dataset (\ie, Kuaishou). 
Amazon is a product dataset with product metadata and corresponding connections. YouTube consists of five types of interactions of YouTube users. 
IMDb is an online dataset of films, videos and television programs.
Taobao is a dataset that contains four types of user behaviors on the online e-commercial platform, Taobao.
Kuaishou dataset consists of three node types and four edge types, which describes four kinds of interactions among users, videos and authors. The statistics of the datasets are summarized in Table \ref{tab:datasets}.
\begin{itemize}
    \item \textbf{Amazon}. The Amazon dataset has a total of $10,099$ nodes and $148, 659$ edges in \textit{Electronics} category. Note that $\mathcal{O}=\{$\textit{product}$\}$, \ie, \textit{I} in the scheme $\mathcal{P}$ and $\mathcal{R}=\{$\textit{common bought}, \textit{common viewed}$\}$.
    \item \textbf{YouTube}. The YouTube dataset has a total of $2,000$ nodes and $1,310,544$ edges. Note that $\mathcal{O}=\{$\textit{product}$\}$, \ie, \textit{I} in the scheme $\mathcal{P}$, and $\mathcal{R}=\{$\textit{contact}, \textit{shared friends}, \textit{shared subscription}, \textit{shared subscriber}, \textit{shared videos}$\}$.
    \item \textbf{IMDb}. The IMDb dataset has $11,616$ nodes and $34,212$ edges. Note that $\mathcal{O}=\{movie, director, actor\}$, \ie, \textit{M, D, A} in the schemes $\mathcal{P}$, respectively.
    \item \textbf{Taobao}. The Taobao dataset describes the interaction between users and items under four different behaviors. Namely, $\mathcal{O}=\{$\textit{user}, \textit{item}$\}$, \ie, \textit{U, I} in the schemes $\mathcal{P}$, respectively, and $\mathcal{R}=\{$\textit{page view}, \textit{add to cart}, \textit{purchase}, \textit{item favoring}$\}$.
    \item \textbf{Kuaishou}. The Kuaishou dataset collects information from the online short-video platform Kuaishou, one of the leading short-video platforms in China (https://www.kuaishou.com/en). We sample user actions among three types of nodes, which are \textit{author}, \textit{user}, \textit{video} (\ie, \textit{A, U, V} in the schemes $\mathcal{P}$, respectively) under four relationships (\ie, \textit{click}, \textit{like}, \textit{comment} and \textit{download}). We sample the data from one-day log, with a total of 105,749 nodes and 175,870 edges. 
\end{itemize}

\begin{table}[t]
\centering
\caption{\todo{The network types handled by different baselines.}}
\setlength{\tabcolsep}{0.8mm}{
\begin{tabular}{c|c|c|c|c}
\toprule
\multirow{2}{*}{Baseline Type}  & \multirow{2}{*}{Baseline}& \multicolumn{2}{c|}{Heterogeneity} & \multirow{2}{*}{Multiplexity} \\
\cline{3-4}
& & Node Type & Edge Type & \\
\midrule
\multirow{3}{*}{\tabincell{c}{Network Embedding\\ Methods}} & DeepWalk & \multirow{3}{*}{Single} & \multirow{3}{*}{Single} & \multirow{3}{*}{False} \\
& node2vec & & \\
& LINE & & \\
\midrule
\multirow{2}{*}{\tabincell{c}{ Homogeneous Graph \\ Neural Networks}} & GCN & \multirow{2}{*}{Single} & \multirow{2}{*}{Single} & \multirow{2}{*}{False} \\
& GraphSage &\\
\midrule
\multirow{2}{*}{\tabincell{c}{Heterogeneous Graph \\ Neural Networks}} 
& HAN & \multirow{2}{*}{Multiple} & \multirow{2}{*}{Multiple} & \multirow{2}{*}{False}\\
& MAGNN &  \\
\midrule
\multirow{2}{*}{\tabincell{c}{Multiplex Heterogeneous \\Graph Neural Networks}} & 
R-GCN & \multirow{2}{*}{Multiple} & \multirow{2}{*}{Multiple} & \multirow{2}{*}{True} \\ 
& GATNE & \\
\bottomrule
\end{tabular}}
\label{tab:baselines_intro}
\end{table}

\subsection{Baselines}
We compare \textit{HybridGNN} with nine state-of-the-art models, including three network embedding models, \ie, DeepWalk \cite{perozzi2014deepwalk}, node2vec \cite{grover2016node2vec}, 
LINE \cite{tang2015line}, two homogeneous graph neural networks, \ie, GraphSage \cite{hamilton2017inductive}, GCN \cite{kipf2017semi}, two heterogeneous graph neural networks, \ie,  HAN \cite{wang2019heterogeneous}, MAGNN \cite{fu2020magnn}, and two multiplex heterogeneous graph neural networks, \ie, R-GCN \cite{schlichtkrull2018modeling} and GATNE \cite{cen2019representation}.
\todo{The heterogeneity and multiplexity of the network types handled by different baselines are listed in Table \ref{tab:baselines_intro}, and the detailed description is as follows. 
Note that we report the best performances of all baselines under best parameter settings in their original papers.}

\begin{table*}
\centering
\caption{The experimental results (\%) for link prediction in the multiplex heterogeneous graph of Amazon, YouTube ($|\mathcal{O}|=1$ and $|\mathcal{R}|\ge 2$) and IMDb ($|\mathcal{O}|\ge 2$ and $|\mathcal{R}|=1$). The best results are illustrated in bold. Please note that the number with a star indicates the result is statistically evaluated with $p<0.01$ under t-test compared to other baselines and the number underlined is the runner-up.}

\resizebox{\textwidth}{!}{
\begin{tabular}{c|ccccc|ccccc|ccccc}
\toprule
& \multicolumn{5}{c|}{Amazon} & \multicolumn{5}{c|}{YouTube} & \multicolumn{5}{c}{IMDb}\\
  & ROC-AUC & PR-AUC & F1 & \todo{PR@10} & \todo{HR@10} & ROC-AUC & PR-AUC & F1 & \todo{PR@10} & \todo{HR@10} & ROC-AUC & PR-AUC & F1 & \todo{PR@10} & \todo{HR@10} \\ 
\midrule
DeepWalk & 95.89 & 95.42 & 90.54 & \todo{0.0096} & \todo{0.0436} & 74.33 & 68.94 & 68.10 & \todo{0.0348} & \todo{0.0118} & 86.47 & 87.10 & 79.54 & \todo{0.0018} & \todo{0.0125} \\
node2vec & 95.16 & 94.13 & 89.34 & \todo{0.0094} & \todo{0.0423} & 77.14 & 72.13 & 70.75 & \todo{0.0404} & \todo{0.0159} & 87.53 & 90.21 & 78.18 & \todo{0.0017} & \todo{0.0114} \\
LINE & 91.71 & 91.82 & 92.01 & \todo{0.0096} & \todo{0.0407} & 76.91 & 71.17 & 70.22 & \todo{0.0403} & \todo{0.0150} & 85.29 & 84.79 & 78.32 & \todo{0.0020} & \todo{0.0135} \\
\midrule
\textit{GCN} & 95.43 & 94.19 & 90.15 & \todo{0.0003} & \todo{0.0014} & 78.01 & 76.86 & 71.26 & \todo{0.0061} & \todo{0.0015} & 87.05 & 90.54 & 79.62 & \todo{0.0004} & \todo{0.0034} \\
\textit{GraphSage} & 96.71 & 96.05 & 91.58 & \todo{0.0044} & \todo{0.0201} & 76.20 & 70.24 & 69.74 & \todo{0.0155} & \todo{0.0052} & 88.07 & 91.32 & 81.27 & \todo{0.0021} & \todo{0.0198} \\
\midrule

HAN & 96.78 & 96.62 & 92.04 & \todo{0.0171} & \todo{0.0561} & 78.36 & 72.74 & 71.26 & \todo{0.0154} & \todo{0.0027} & \underline{89.44} & 92.01 & 82.75 & \todo{0.0248} & \todo{0.2221} \\
MAGNN & 96.99 & 96.48 & 91.94 & \todo{0.0118} & \todo{0.0357} & 79.75 & 75.03 & 72.53 & \todo{0.0369} & \todo{0.0028} & 88.87 & 91.75 & 81.46 & \todo{0.0638} & \todo{0.5125} \\
\midrule
\todo{R-GCN} & \todo{97.26} & \todo{96.07} & \todo{\underline{93.12}} & \todo{0.0318} & \todo{0.1137} & \todo{80.60} & \todo{75.31} & \todo{72.98} & \todo{0.0367} & \todo{0.0133} & \todo{87.46} & \todo{88.89} & \todo{82.59} & \todo{0.0468} & \todo{0.3932} \\
GATNE & \underline{97.44} & \underline{97.05} & 92.87 & \todo{\underline{0.0392}} & \todo{\underline{0.1440}} & \underline{84.61} & \underline{81.93} & \underline{76.83} & \todo{\underline{0.0435}} & \todo{\underline{0.0258}} & 89.22 &  \underline{93.02} & \underline{83.12} & \todo{\underline{0.0820}} & \todo{\underline{0.6192}} \\

\midrule
\textit{HybridGNN} & \textbf{97.79}$^*$ & \textbf{97.47}$^*$ & \textbf{93.51}$^*$ & \todo{\textbf{0.0430}} & \todo{\textbf{0.1613}} & \textbf{86.22}$^*$ & \textbf{85.16}$^*$ & \textbf{79.07}$^*$ & \todo{\textbf{0.0461}} & \todo{\textbf{0.0264}} & \textbf{90.94}$^*$ & \textbf{93.44}$^*$ & \textbf{84.26}$^*$  & \todo{\textbf{0.1074}} & \todo{\textbf{0.7684}} \\
\bottomrule
\end{tabular}}

\label{tab:baselines1}
\end{table*}

\begin{table*}
\centering
\caption{The experimental results (\%) for link prediction in the multiplex heterogeneous graph of Taobao and Kuaishou dataset ($|\mathcal{O}|\ge 2$ and $|\mathcal{R}|\ge 2$). The best results are illustrated in bold. Please note that the number with a star indicates the result is statistically evaluated with $p<0.01$ under t-test compared to other baselines and the number underlined is the runner-up.}
\resizebox{!}{2.1cm}{
\begin{tabular}{c|ccccc|ccccc}
\toprule
&  \multicolumn{5}{c|}{Taobao} & \multicolumn{5}{c}{Kuaishou}\\
  & ROC-AUC & PR-AUC & F1 & \todo{PR@10} & \todo{HR@10} & ROC-AUC & PR-AUC & F1 & \todo{PR@10} & \todo{HR@10} \\ 
\midrule
DeepWalk & 88.21 & 87.98 & 80.39 & \todo{0.0102} & \todo{0.0944} & 86.93 & 83.53 & 73.24 & \todo{0.0043} & \todo{0.0420} \\
node2vec & 88.02 & 87.60 & 80.24 & \todo{0.0091} & \todo{0.0841} & 85.93 & 82.49 & 70.82 & \todo{0.0035} & \todo{0.0345}\\
LINE &  87.68 & 90.39 & 79.59 & \todo{0.0099} & \todo{0.0928} & 86.99 & 83.59 & 73.40 & \todo{0.0048} & \todo{0.0445} \\
\midrule
\textit{GCN} & 91.12 & 92.38 & 83.07 & \todo{0.0002} & \todo{0.0019} & 87.66 & 84.68 & 74.38 & \todo{0.0018} & \todo{0.0131}\\
\textit{GraphSage} & 92.90 & 93.12 & 84.99 & \todo{0.0009} & \todo{0.0036} & 87.02 & 83.70 & 72.02 & \todo{0.0104} & \todo{0.0889} \\
\midrule

HAN & 93.00 & 93.13 & 84.89 & \todo{0.0025} & \todo{0.0200} & 88.46 & 86.35 & 76.31 & \todo{0.0077} & \todo{0.0730} \\
MAGNN & 95.26 &	95.61 &	88.52 & \todo{0.0130} & \todo{0.0857} & 89.11 & 87.15 & 77.43 & \todo{0.0234} & \todo{0.2067}\\
\midrule
\todo{R-GCN} & \todo{96.59} & \todo{95.29} & \todo{91.34} & \todo{0.0123} & \todo{0.1148} & \todo{86.75} & \todo{87.09} & \todo{78.44} & \todo{0.0212} & \todo{0.1803} \\
GATNE & \underline{97.19} & \underline{97.82} & \underline{92.53} & \todo{\underline{0.0214}} & \todo{\underline{0.1175}} & \underline{91.83} & \underline{91.32} & \underline{82.72} & \todo{\underline{0.0393}} & \todo{\underline{0.3344}}\\

\midrule
\textit{HybridGNN} & \textbf{98.45}$^*$ & \textbf{98.77}$^*$ & \textbf{95.61}$^*$ & \todo{\textbf{0.0217}} & \todo{\textbf{0.1281}} & \textbf{92.11*} & \textbf{92.50*} & \textbf{86.02*} & \todo{\textbf{0.0430}} & \todo{\textbf{0.3911}} \\
\bottomrule
\end{tabular}
\label{tab:baselines2}
}
\end{table*}

\subsubsection{Network Embedding Methods}
\begin{itemize}
    \item \textbf{DeepWalk} \cite{perozzi2014deepwalk} is a graph embedding method for homogeneous networks. It uses a random walk strategy to sample a series of nodes and then trains a skip-gram model. We ignore the node types and edge types in the heterogeneous network when training and evaluating DeepWalk.
    \item \textbf{node2vec} \cite{grover2016node2vec} learns the node embedding through an advanced sampling strategy from DeepWalk. We apply node2vec using the same way as DeepWalk.
    \item \textbf{LINE} \cite{tang2015line} is a graph embedding method for homogeneous networks, which learns the node embedding through first-order and second-order proximities between two nodes. We train LINE by hiding the node types and edge types.
\end{itemize}
\subsubsection{Homogeneous Graph Neural Networks}
\begin{itemize}
    \item \textbf{GCN} \cite{kipf2017semi} is a graph convolutional network for the homogeneous network. The heterogeneous information is ignored during training and testing GCN. 
    \item \textbf{GraphSage} \cite{hamilton2017inductive} is a graph neural network for both transductive and inductive representation learning in homogeneous networks. We ignore the heterogeneity in the networks when we evaluate the performance of GraphSage.
\end{itemize}  

\subsubsection{Heterogeneous Graph Neural Networks}
\begin{itemize}
    \item \textbf{HAN} \cite{wang2019heterogeneous} is a heterogeneous graph attention network. The node embeddings are generated through a hierarchical attention module. As it ignores the multiplexity property in the heterogeneous network, we report its best performance in a non-multiplex heterogeneous network among all the metapaths' candidates.
    \item \textbf{MAGNN} \cite{fu2020magnn} employs the metapath aggregation to learn the node embedding in the non-multiplex heterogeneous network. We also report its best performance among all the metapaths' candidates. 
\end{itemize}

\subsubsection{Multiplex Heterogeneous Graph Neural Networks}
\begin{itemize}
    \item \todo{\textbf{R-GCN}}\todo{\cite{schlichtkrull2018modeling} is a relational graph convolutional network. It uses an encoder to learn feature representations and a decoder to predict labels. We use the triplet scoring function and the corresponding cross-entropy objectives.}
    \item \textbf{GATNE} \cite{cen2019representation} is a node embedding learning method for multiplex heterogeneous networks.
\end{itemize}

\subsection{Experimental Settings}
We use grid search to find the optimal hyper-parameters for \textit{HybridGNN}. 
Specifically, we set the basic embedding $e_{v_i}$ as $[64, 128, 256, 512]$, the local edge embedding $e_{v_i, r_l}$ as $[2, 8, 64, 128]$ and the negative node numbers as $[1,3,5,7]$. 
Moreover, the batch size is $2048$ for all the datasets. 
We use Adam optimizer to train our model with an initial learning rate $5e-3$ in Amazon, YouTube, IMDb, and $5e-4$ in Taobao and Kuaishou. 
For all the datasets, we split $85\%$ edges as training edges, $5\%$ edges as validation edges and the rest $10\%$ edges as test edges. 
In the validation and test set, we also randomly sample the same number of negative edges once a positive edge is generated. 
The strategy of early stopping is adopted and the whole training procedure is stopped if there is no further improvement in the following five epochs. 
For sampling based methods that adopt random walks to generate positive neighbors, the number of walks is set to 20 and the length of walk is set to 10. The window size is set to 5. We sample 5 negative neighbors for each training pair.

The experiments conducted on Amazon, YouTube, IMDb and Taobao datasets are trained on a single GTX 1080Ti GPU, and the codes used for training and evaluation are implemented with PyTorch 1.7.0 in Python 3.6. For Kuaishou dataset, it is trained on 8 GTX 1080Ti GPUs, the codes are implemented with PyTorch 1.1.0 in Python 3.6.


Following \cite{cen2019representation}, we choose three criteria to evaluate link prediction results. Evaluation metrics are ROC-AUC \cite{hanley1982meaning}, PR-AUC \cite{davis2006relationship} and F1 scores. \todo{Besides, for nodes in the testing set, we evaluate the top-$K$ recommendation performance and report the average PR@$K$(Precision) and HR@$K$(Hit Ratio) where $K=10$.}

\begin{table*}[t]
\centering
\caption{Performances of different search scopes of randomized inter-relationship exploration. Note that $L$ represents the depth of randomized exploration.}
\begin{tabular}{c|cc|cc|cc|cc}
\toprule
Depth & \multicolumn{2}{c|}{Amazon} & \multicolumn{2}{c|}{YouTube} & \multicolumn{2}{c|}{IMDb} & \multicolumn{2}{c}{Taobao} \\
\midrule 
\textit{HybridGNN} ($L=1$) & $\textbf{97.72}$ & $\textbf{93.36}$ & $85.26$ & $78.13$ & $89.54$ & $83.39$ & 98.24 & $94.85$ \\
  \textit{HybridGNN} ($L=2$) & $97.67$ & $93.33$ & $\textbf{85.67}$ & $78.64$ & $\textbf{89.78}$ & $\textbf{83.60}$ & $\textbf{98.64}$ & $\textbf{95.81}$ \\
\textit{HybridGNN} ($L=3$) & $97.65$ & $93.32$ & $85.64$ & $\textbf{78.70}$ & $89.72$ & $83.49$ & $98.01$ & $94.39$ \\
\bottomrule
\end{tabular}
\label{tab:diff_randomzied}
\end{table*}

\subsection{Link Prediction Task}
To evaluate the performance of  \textit{HybridGNN}, we compare our model with other baselines under different types of heterogeneous networks, \ie, $\mathcal{G}_1$ with $|\mathcal{O}|=1$ and $|\mathcal{R}|\ge 2$, $\mathcal{G}_2$ with $|\mathcal{O}|\ge2$ and $|\mathcal{R}|= 1$, $\mathcal{G}_3$ with $|\mathcal{O}|\ge 2$ and $|\mathcal{R}|\ge 2$. We report the overall results in Table \ref{tab:baselines1} and Table \ref{tab:baselines2}.

\textbf{Comparison with homogeneous graph models}. 
We first compare the results with network embedding methods such as network embedding methods (i.e., DeepWalk, node2vec and LINE) and homogeneous graph neural networks (i.e., GCN and GraphSage). 
As is shown in Table \ref{tab:baselines1} and Table \ref{tab:baselines2}, although GNNs aggregate information from neighbors, the aggregated features fail to improve the performances on all the datasets. 
One possible reason is that the heterogeneity in HINs prefers a selective aggregation strategy, which causes the random aggregation in traditional GNNs fails to approximate the best performances constantly. \textit{HybridGNN} outperforms these homogeneous baselines with $3\%\sim 8\%$. 

\textbf{Comparison with heterogeneous graph models}. 
We then compare the results with the state-of-the-art models \todo{(\ie,  HAN, MAGNN, RGCN and GATNE)} for heterogeneous networks. 
We observe that heterogeneous graph neural networks achieve a better performance than network embedding methods, with $0.7\%\sim 2.1\%$ uplift of ROC-AUC, PR-AUC and F1 on different datasets. \todo{The improvement on top-$K$ recommendation metric is more significant on datasets with more distinct multiplex characteristics like Taobao and Kuaishou.}
These results indicate the importance of explicitly modeling the heterogeneity in HINs. 
Moreover, models with consideration of multiplexity property in HINs achieve higher performances. 
As non-multiplex aware methods only learn one node embedding among relationships, the diversity is not well captured. \textit{HybridGNN} outperforms both multiplex aware methods and non-multiplex aware methods, suggesting the effectiveness of the discovery mechanism on approximation towards the inter-relationships.
\begin{table}
\centering
\begin{threeparttable}[b]
\caption{Uplift from randomized inter-relationship exploration. Note that the performances are evaluated under AUC-ROC and evaluation metrics are reported under a specific relationship.}
\begin{tabular}{c|c|c|c}
\toprule
 $g\subseteq\mathcal{G}$ & GCN & GATNE & \textit{HybridGNN}  \\
\midrule
 $g_{r_0}$ & 80.63 & 82.92 & \textbf{82.97}  \\
 $g_{r_0, r_1}$ & 80.63 & 84.17 & \textbf{86.60}  \\
 $g_{r_0, r_1, r_2}$ & 80.63 & 84.37 & \textbf{87.05}  \\
 $g_{r_0, r_1, r_2, r_3}$ & 80.63 & 87.01 & \textbf{87.82} \\
 $g_{r_0, r_1, r_2, r_3, r_4}$ & 80.63 & 88.04 & \textbf{88.73}  \\
\bottomrule
\end{tabular}
\label{tab:relational_comp}
\end{threeparttable}
\end{table}

\subsection{Analysis of Randomized Exploration}
We conduct experiments to evaluate the performance improved by the randomized inter-relationship exploration approach. 
To have a comprehensive understanding of our randomized exploration module, in this subsection, we compare different search scopes to analyze whether a deeper randomized exploration can help to bring positive benefits to the \textit{HybridGNN} model.


Table \ref{tab:diff_randomzied} shows that, in general, a deeper randomized exploration search scope cannot always help the improvement. For example, the ROC-AUC and F1 scores decrease on the Amazon dataset when the depth of randomized exploration increases from $1$ to $3$. This phenomenon may be related to the complexity of the multiplex heterogeneous network. If $\mathcal{G}$ has abundant node types and edge types (relationships), the latent semantic metapath schemes can be far more than the relatively simple multiplex heterogeneous networks under the same condition. For instance, the IMDb dataset has three node types and one relationship, and Taobao has two node types and four relationships. The best performances on these datasets increase with reasonably deeper randomized aggregation layer.
However, the number of meaningless metapath schemes (\ie, noises) also grows in the searching scope with the randomized aggregation layer deepening unlimitedly.
There is a peak of choosing the best aggregation layer with different datasets. Basically, \textit{HybridGNN} achieves the best performance in $L=2$ when $\mathcal{G}$ is complex.


\begin{table}[t]
\centering
\caption{Quantitative results (F1 score) for ablation study.}
\begin{tabular}{c|c|c|c|c}
\toprule
Models & Amazon & YouTube & IMDb & Taobao \\
\midrule 
\textit{HybridGNN} & \textbf{93.51} & \textbf{79.07} & \textbf{84.26} & \textbf{95.61} \\
\textit{w/o} metapath-level attention & 93.29 & 78.14 & 83.37 & 93.25 \\
\textit{w/o} relationship-level attention & 93.40 & 78.62 & 83.55 & 91.64  \\
\textit{w/o} randomized exploration & 93.45 & 77.92 & 83.43 & 89.45 \\
\textit{w/o} hybrid aggregation flow & 93.41 & 76.42 & 83.12 & 89.02 \\
\bottomrule
\end{tabular}
\label{tab:quantitative}
\end{table}

\subsection{Analysis of Uplift from Inter-relationship}
We conduct the experiments to evaluate the uplifts from the incorporation of inter-relationship information.
The experimental settings are as follows: We firstly extract a subgraph $g_{r_i}$ (\ie, $g_{r_0}$) from the YouTube dataset to get the knowledge on the performance of the model only seeing $g_{r_i}$. Then, as the subgraph is enlarged with another relationship $r_j$ added, we test the performance under $g_{r_i, r_j}$. This procedure is repeated until $g_{r_i, r_j, \cdots, r_k}$ becomes $\mathcal{G}$. Thus, we can analyze the benefits of introducing inter-relationship information to \textit{HybridGNN}.

Table \ref{tab:relational_comp} shows that the performances of multiplex heterogeneous graph neural networks under a specific relationship improve with other relationship-specific subgraphs included. 
As GCN is a homogeneous graph neural network, enlarging subgraph from $g_{r_i}$ to $\mathcal{G}$ cannot improve the model performance, and the results are consistent. Consequently, \textit{HybridGNN} outperforms GATNE under each subgraph in terms of AUC-ROC. It indicates the effectiveness of the randomized inter-relationship exploration.

\subsection{Ablation Study}
In this subsection, we conduct the ablation study of \textit{HybridGNN}. Specifically, there are four variants. We denote \textit{HybridGNN} \textit{w/o} metapath-level as the removal of metapath-level attention, \textit{HybridGNN} \textit{w/o} relationship-level as the removal of relationship-level attention,
\textit{HybridGNN} \textit{w/o} randomized as the removal of randomized inter-relationship exploration (\ie, the model only aggregates intra-relationship metapaths) in \textit{HybridGNN} and 
\textit{HybridGNN} \textit{w/o} hybrid as the replacement of hybrid aggregation flows with random sampling aggregation.

As shown in Table \ref{tab:quantitative}, hierarchical attention, randomized inter-relationship exploration and hybrid aggregation flows take contribution differently to the performance of \textit{HybridGNN}. Generally, the removal of randomized exploration and the replacement of hybrid aggregation flows have the most important influence on \textit{HybridGNN}. There are $1\%\sim3\%$ gap on F1 scores compared to the other two variants on YouTube, IMDb and Taobao datasets, and the removal of hierarchical attention has the most important influence on Amazon.

\begin{figure}[t]
  \centering
  \subfigure[Impact of $d_m$]{
    \begin{minipage}[]{.42\linewidth}
    \includegraphics[width=1.0\linewidth]{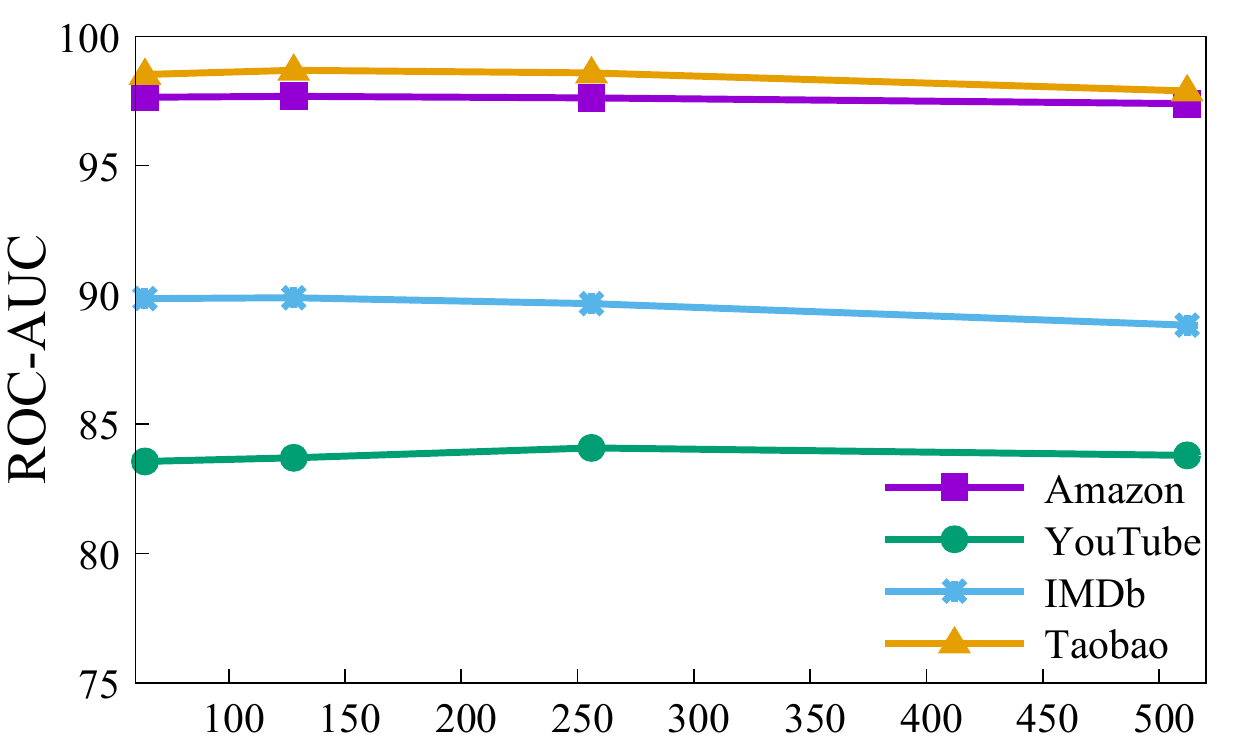}
    \end{minipage}
  }
  \subfigure[Impact of $d_e$]{
    \begin{minipage}[]{.42\linewidth}
   \includegraphics[width=1.0\linewidth]{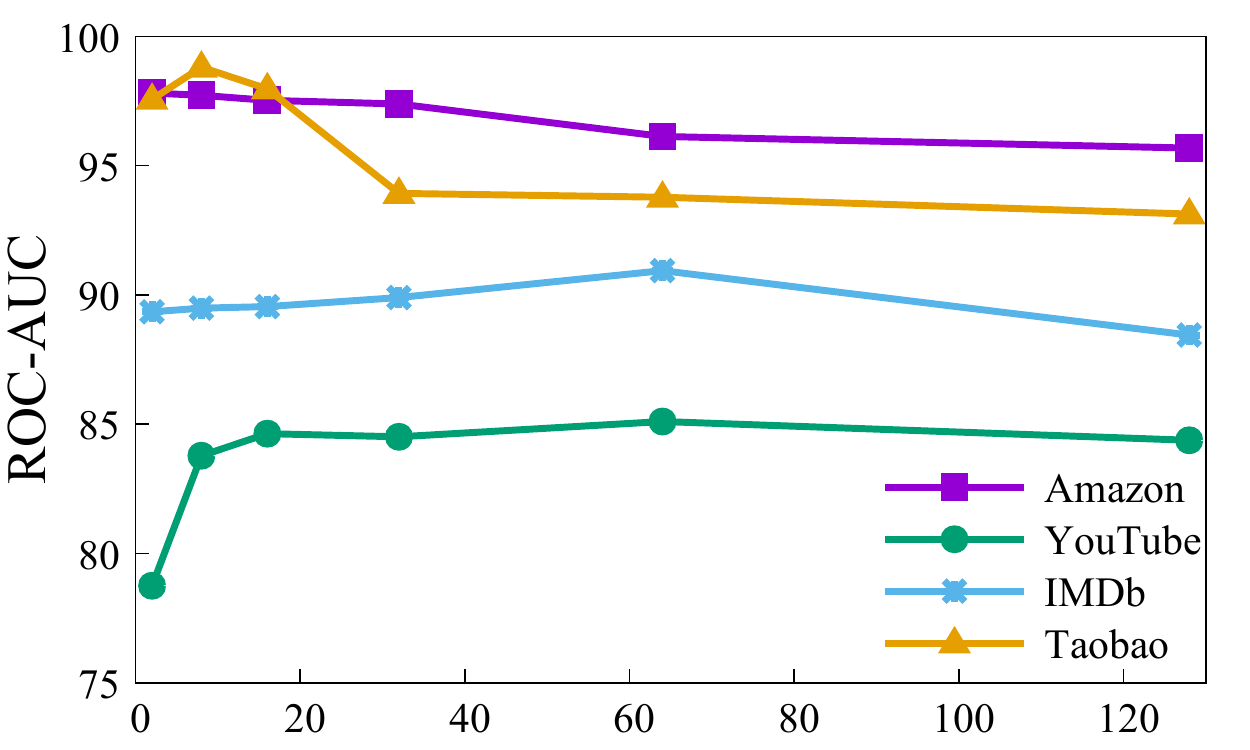}
    \end{minipage}
  }
  \subfigure[Impact of $n$]{
    \begin{minipage}[]{.42\linewidth}
   \includegraphics[width=1.0\linewidth]{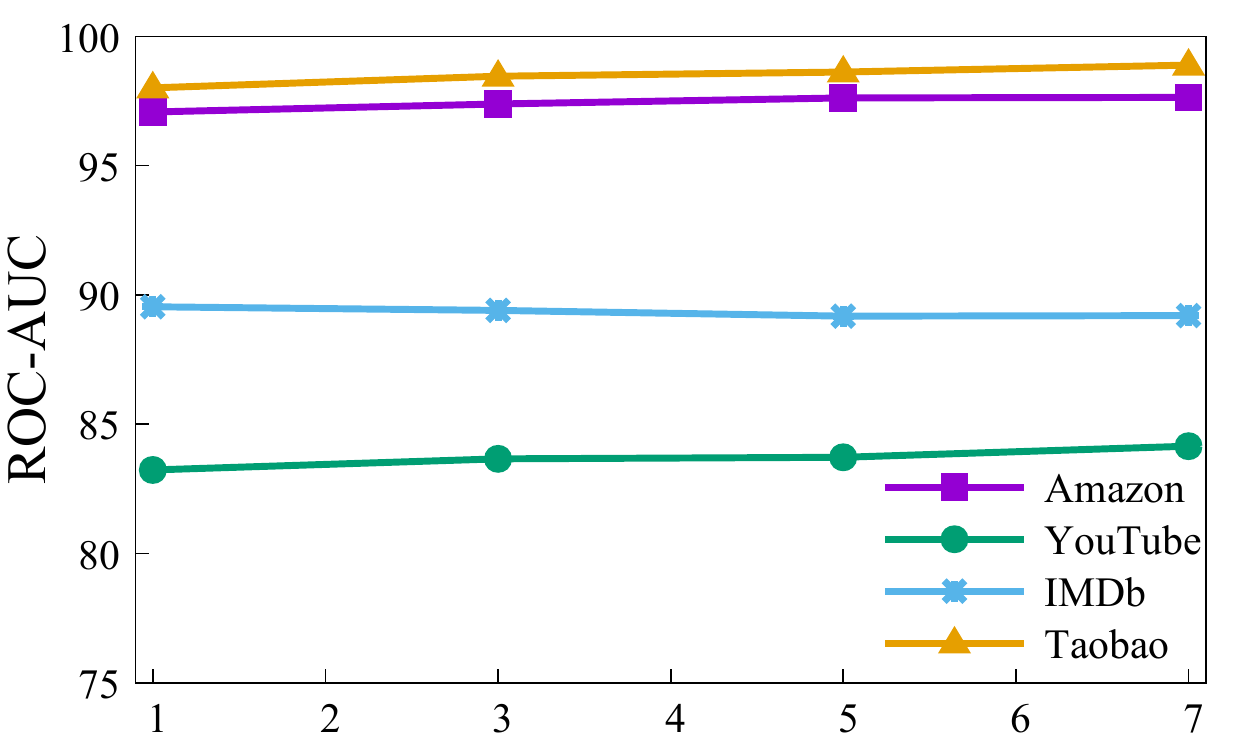}
    \end{minipage}
  }
   \caption{Impacts of hyper-parameters on different datasets.}\label{fig:parameter_sensitiviy}
\end{figure}
\subsection{Parameter Sensitivity}
The hyper-parameter sensitivity of \textit{HybridGNN} is analyzed on four datasets. 
Specifically, the impact by the number of negative nodes $n$, the dimension of the basic node embedding $d_m$ and that of the edge embedding $d_e$ is reported. 

As shown in Fig.~\ref{fig:parameter_sensitiviy}, the performance of \textit{HybridGNN} is relatively smooth with $d_m$ ranging inside $[64, 128, 256, 512]$. Moreover, it indicates \textit{HybridGNN} achieves the best performance when $d_m=128$ on all the datasets.
In addition, \textit{HybridGNN} is stable with $d_e$ ranging over $[2, 8, 16, 32, 64, 128]$. Also, it shows that $d_e=8$ may be a better choice.
For the negative node number $n$, the performance of \textit{HybridGNN} slightly increases when $n$ increases from $1$ to $5$.

\begin{figure}[t]
  \centering
  \subfigure[\todo{Taobao}]{
    \begin{minipage}[]{.45\linewidth}
    \includegraphics[width=1.0\linewidth]{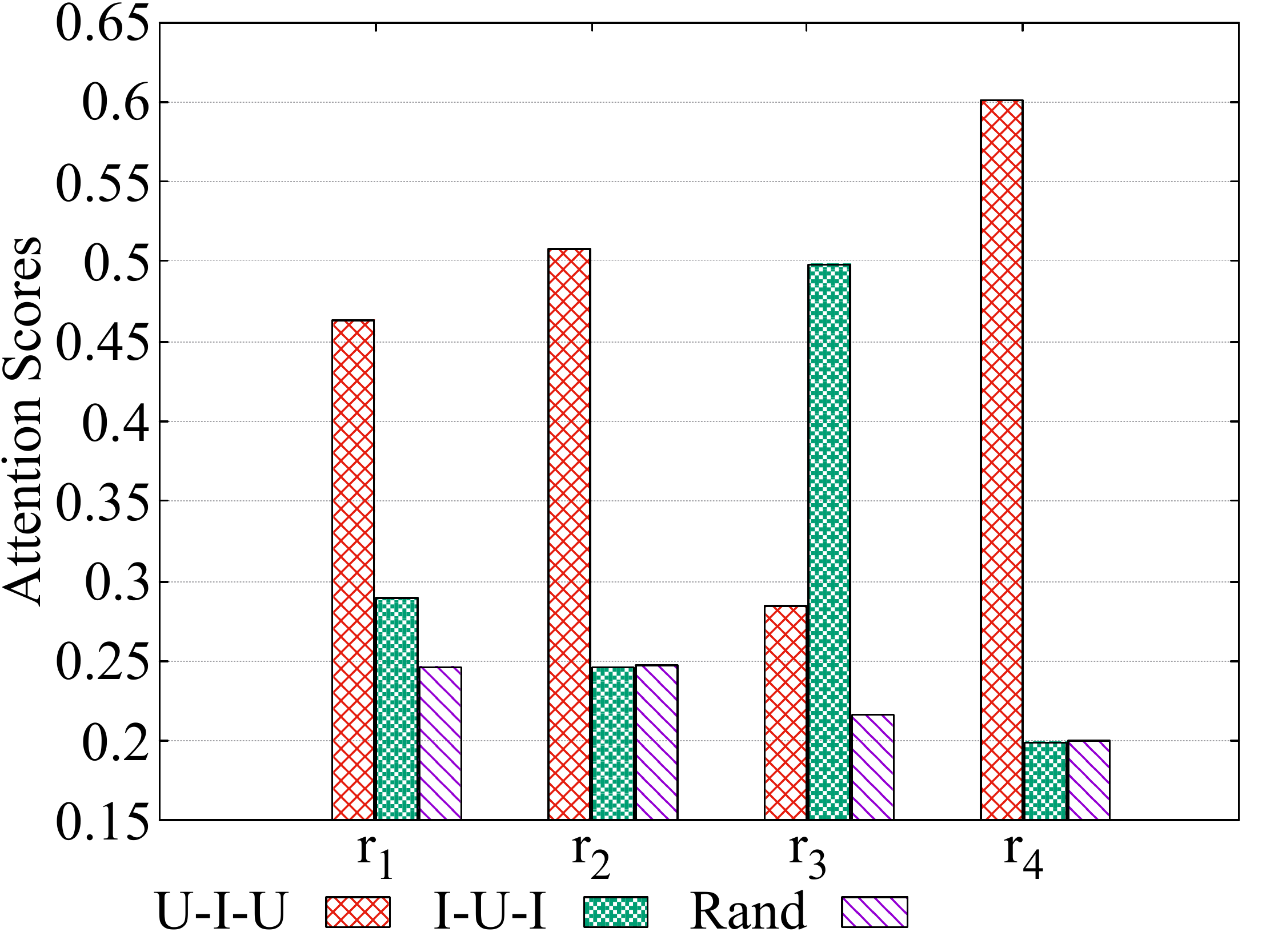}
    \end{minipage}
  }
  \subfigure[\todo{Kuaishou}]{
    \begin{minipage}[]{.45\linewidth}
  \includegraphics[width=1.0\linewidth]{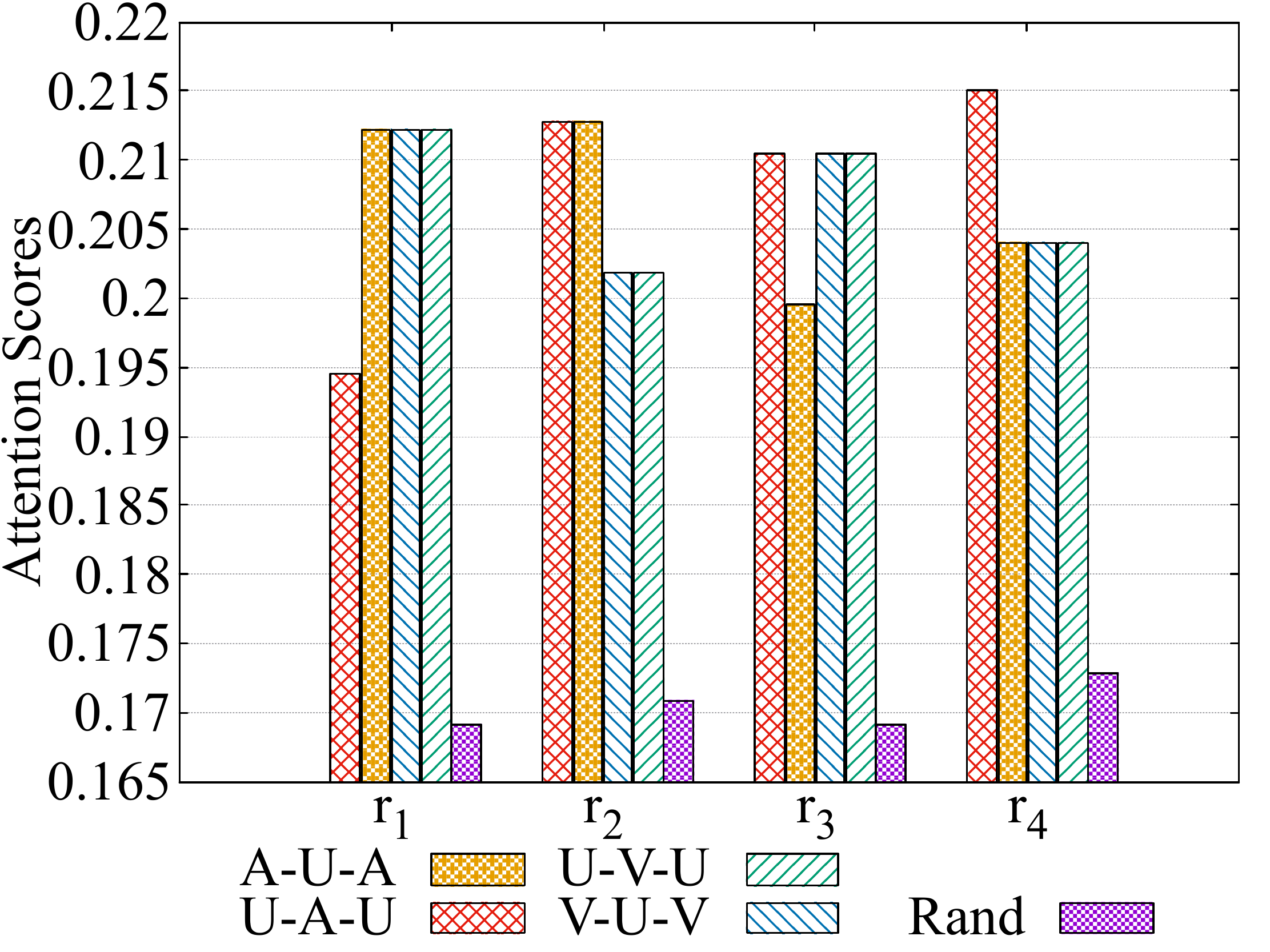}
    \end{minipage}
  }
  \caption{\todo{Attention scores of different metapaths under each relationship on Taobao and Kuaishou datasets. Note that $r_1$, $r_2$, $r_3$ and $r_4$ in Taobao are \textit{page view}, \textit{item favoring}, \textit{purchase} and \textit{add to cart}, respectively, while those in Kuaishou are \textit{click}, \textit{like}, \textit{comment} and \textit{download}, respectively.}}\label{fig:case_study_att_dis}
\end{figure}

\begin{figure}[t]
    \centering
    \includegraphics[width=0.6\linewidth]{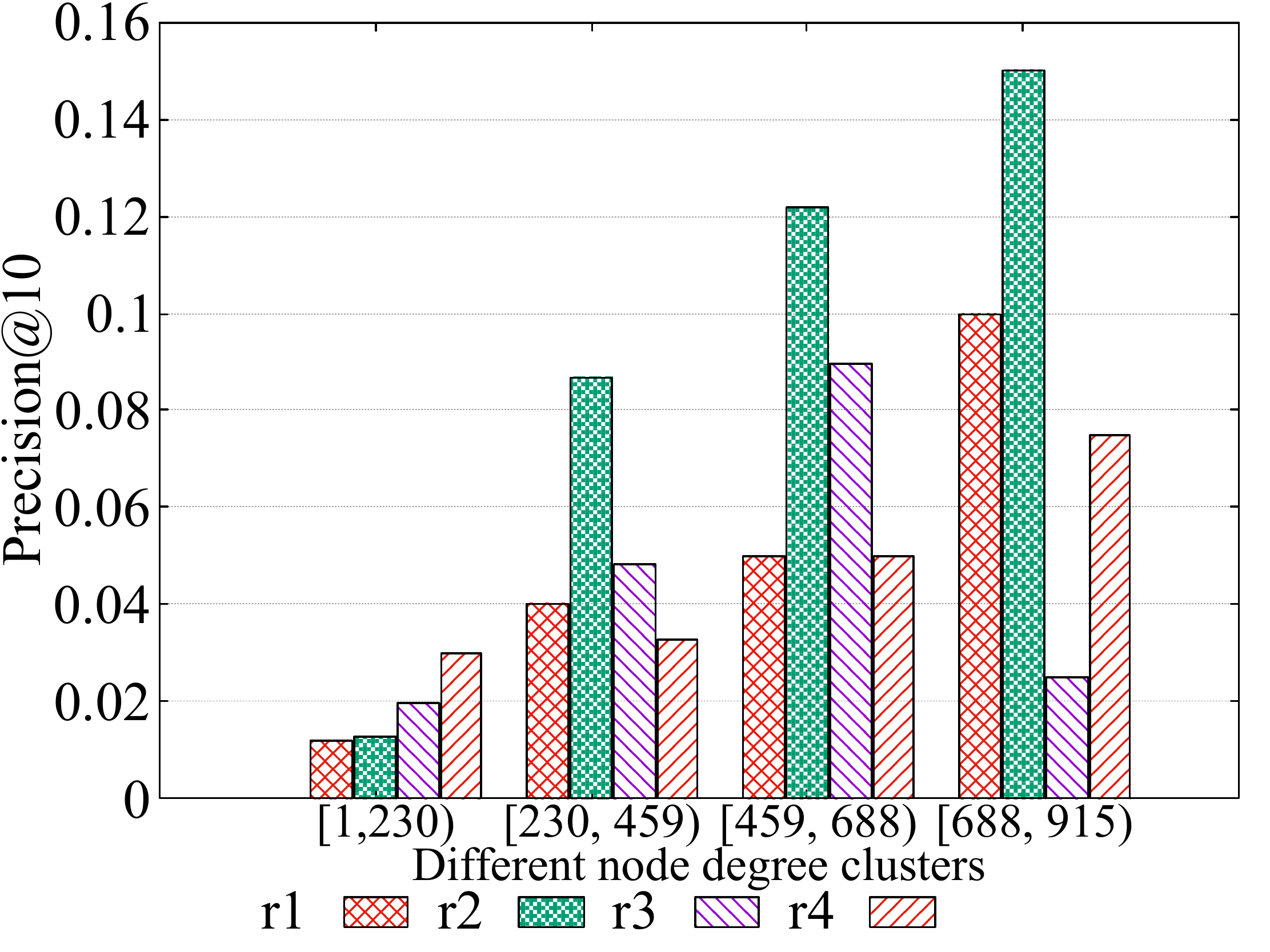}
    \caption{\todo{The performance of node clusters of different degrees on the Taobao dataset. Note that $r_1$, $r_2$, $r_3$ and $r_4$ are \textit{page view}, \textit{item favoring}, \textit{purchase} and \textit{add to cart}, respectively.}}\label{fig:degree_tb}
\end{figure}

\begin{table}[t]
\centering
\caption{\todo{The performance comparison with GATNE of different degree clusters in terms of PR@$K$ on the IMDb dataset. Note that $d_n$ denotes the node degree.}}
\begin{tabular}{c|c|c|c|c}
\toprule
&  1$\leq$$d_n$\textless 20 & 20$\leq$$d_n$\textless 39 & 39$\leq$$d_n$\textless 58 & 58$\leq$$d_n$\textless 76 \\
\midrule 
\textit{GATNE} & 0.1044 & 0.1699 & 0.2095 & 0.1000 \\
\textit{HybridGNN} & \textbf{0.1054} & \textbf{0.1880} & \textbf{0.2714} & \textbf{0.1500} \\
\midrule
\textit{Improvement} & 0.96\% & 10.84\% & 29.55\% & 50.00\%  \\
\bottomrule
\end{tabular}
\label{tab:degree_comp}
\end{table}
\subsection{Case study}
\todo{We conduct a case study, aiming to figure out the following questions: 
a) How does the attention distributions of different metapaths vary with different types of relationships in the multiplex scenarios? 
b) How does the recommendation performance vary with nodes of different degrees?}

\todo{We firstly analyze the attention scores of different metapaths under each relationship on Taobao and Kuaishou datasets, respectively.
As shown in Fig.~\ref{fig:case_study_att_dis}, the impact of different metapaths varies with different relationships. 
Although the metapath scheme \textit{U-I-U} has a large impact in most of the relationships on the Taobao dataset (\ie, $r_1, r_2, r_4$ in Fig.~\ref{fig:case_study_att_dis}(a)), the attention scores of the other paths vary. 
For instance, in $\mathcal{G}_{r_4}$, the random aggregation flow takes little effect as $\mathcal{G}_{r_4}$ has rich \textit{user-item} interactions. 
In contrast, \textit{user-item} interactions in $\mathcal{G}_{r_1}$ and $\mathcal{G}_{r_2}$ are sparse. 
Therefore, the random aggregation flow contributes to a better node representation learning through \textit{randomized exploration}. 
For the Kuaishou dataset, the metapaths take a comparatively balanced effect while the random sampled path takes an auxiliary effect. 
It illustrates that selected metapaths can cover most of the rich interactions, and the remaining useful information is limited.}



\todo{
Further, we analyze the success and failure predictions of \textit{HybridGNN}. Note that the model performance is highly correlated with sampled neighbors of the target node, and the sampling method is greatly influenced by the node's degree, so we study the performance discrepancy of nodes with different degrees.
We use the metric PR@$10$ to represent the confidence level of predictions. 
Fig.~\ref{fig:degree_tb} shows the performance of nodes with different degrees on the Taobao dataset, where we divide the nodes into four clusters based on their degrees.
The results indicate that the recommendation for nodes with higher degrees under different relationships is generally better, and the reason is \textit{HybridGNN} samples richer metapath-guided neighbors for nodes with higher degrees. Besides, we compare our model with \textit{GATNE} of different degree clusters on the IMDb dataset. The results shown in Table \ref{tab:degree_comp} demonstrate the better recommendation capability of \textit{HybridGNN} on nodes with high degrees as well.
}



\section{Related Work}
In this section, we review related work relevant to our study, including network embedding, heterogeneous graph neural networks, and graph neural networks for recommendations.

\label{rw:gra_neu_netw}
\subsection{Network Embedding}
Graph structures are ubiquitous in real-world applications and are natural to model complex interactions. 
Network embedding is to learn low dimensional vectors of nodes or edges with network structure preserved. 
There are many studies on learning homogeneous networks \todo{ \cite{perozzi2014deepwalk, tang2015line, grover2016node2vec, wang2016structural, yan2020tinygnn, wang2020edge2vec}}.
\todo{SDNE\cite{wang2016structural} learns network embedding with an aim to both capture high non-linearity and preserve structural information.}
Kipf et al.~\cite{kipf2017semi} propose a graph convolutional network (GCN) via the approximation of spectral graph convolutions, which incorporates the node features from its neighbors and aggregates information through multi-layer integration. 
Nevertheless, propogation in a single layer of GCN involves overall spatially close neighbors, and is costly with graphs increasing. 
Hamilton et al.~\cite{hamilton2017inductive} propose GraphSage, which aggregates information through a neighborhood sampling function and constrains a fixed sampling number in each layer. 
It is also an inductive framework for learning embeddings of unseen nodes by leveraging node features. 
Graph attention network \cite{velivckovic2017graph} is proposed to further exploit the performance improved by attention mechanisms \cite{bahdanau2015neural}.
Edge2vec\cite{wang2020edge2vec} is a pioneer work to learn edge embeddings in social networks through preserving both the local and the global information.

As the above models are dedicated for homogeneous networks, the rich semantic information, including heterogeneous topology and interactive patterns, is ignored.

\subsection{Heterogeneous Graph Neural Networks}
\label{rw:net_rep_learn}
In heterogeneous networks, representation learning is to embed the whole graph into a low dimensional space with both topological structure and abundant side-information preserved.
Learned embeddings can preserve useful semantics from complex graphs for downstream tasks such as node classification \cite{wang2019heterogeneous}, node clustering \cite{chang2015heterogeneous} and link predictions \cite{cen2019representation}.
A variety of methods are proposed for learning representations in heterogeneous networks \cite{dong2017metapath2vec, fu2017hin2vec}.
\todo{R-GCN \cite{schlichtkrull2018modeling} is firstly proposed to model the relational data under GCN framework for fundamental statistical relational learning tasks.
The link prediction task is conducted in an autoencoder manner. 
The encoder learns the latent feature representations while the decoder aims to predict the labeled edges with factorization method.} 
Wang et al.~\cite{wang2019heterogeneous} attempt to study the heterogeneous information network based on node-level and semantic-level attention.
Nevertheless, the intermediate semantics along metapaths are neglected.
Thereafter, MAGNN \cite{fu2020magnn} is proposed to further consider intermediate nodes for learning metapath information in HINs.
Lai et al.~\cite{lai2020policy} propose a Policy-GNN, an effective aggregation strategy, to determine the iteration of aggregations. 
HetGNN \cite{zhang2019heterogeneous} is a heterogeneous contents aware model for inductive learning.

In this paper, our target is not only to learn representations from complex objects and rich interactions, but also to model the multiplexity in heterogeneous networks, which  fully utilizes multiple edges between two nodes.

\subsection{Graph Neural Networks for Recommendations}
As heterogeneous networks naturally model different types of objects and relationships, recent studies have emerged to exploit the heterogeneity in recommender systems, \todo{such as learning representation from rich interactions \cite{wang2019neural,wang2020disentangled,he2020lightgcn}, learning price-aware recommendations \cite{zheng2020price} and group recommendations \cite{guo2020group, deng2021knowledge} in e-commerce systems}.
CSE \cite{chen2019collaborative} is a unified framework for representation learning. 
Two types of proximity relationships, \ie, direct similarity and neighborhood similarity, are considered.
LightGCN \cite{he2020lightgcn} is designed to simplify GCN with neighborhood aggregations.
To fully incorporate interactive patterns between users and items, Jin et al. \cite{jin2020efficient} introduce NIRec to avoid explicit path reachability with rich semantics preserved in HINs.
Fan et al. \cite{fan2019metapath} also address the importance of learning meaningful semantics from metapaths and propose a metapath-guided embedding method.

However, the above methods usually assume a single relationship between two nodes and thus ignore situations where two nodes can have multiple relationships in real-world applications. 
Thus the multiplexity property in HINs is not fully utilized.
MNE \cite{zhang2018scalable} projects a node representation of different types into a unified embedding space, and uses a common embedding with additional edge embedding to represent a node with different edge types.
GATNE \cite{cen2019representation} further learns heterogeneity in multiplex heterogeneous network based recommender systems with attention mechanisms. 
However, these methods only aggregate neighbors in relationship-specific subgraphs and fail to benefit from the inter-relationship information.



\section{Conclusion}
In this paper, we propose the \textit{HybridGNN} model, tending to capture different behaviors elegantly and efficiently under different relationships in multiplex HINs. \textit{HybridGNN} uses hybrid aggregation flows and hierarchical attentions to fully utilize the heterogeneity in the multiplex scenarios. To explore the importance of different aggregation flows and capture informative messages from other relationships, we propose a novel hierarchical attention module which leverages both metapath-level attention and relationship-level attention. \textit{HybridGNN} achieves significantly better results compared to other state-of-the-art methods for both homogeneous network embedding and heterogeneous network embedding in the link prediction task.
In the future, HybridGNN will be developed to incorporate more side information of multiplex heterogeneous networks.

\section*{Acknowledgments}
This work is supported in part by the National Natural Science Foundation of China (No.~61872207) and Kuaishou Inc. 
Chaokun Wang is the corresponding author.

\bibliographystyle{IEEEtran}
\bibliography{sample-base}

\end{document}